\documentclass[pdflatex,sn-mathphys-num,ay]{sn-jnl}

\usepackage{graphicx}%
\usepackage{multirow}%
\usepackage{amsmath,amssymb,amsfonts}%
\usepackage{amsthm}%
\usepackage{mathrsfs}%
\usepackage[title]{appendix}%
\usepackage{xcolor}%
\usepackage{textcomp}%
\usepackage{manyfoot}%
\usepackage{booktabs}%
\usepackage{algorithm}%
\usepackage{algorithmicx}%
\usepackage{algpseudocode}%
\usepackage{listings}%
\theoremstyle{thmstyleone}%
\theoremstyle{thmstyletwo}%

\theoremstyle{thmstylethree}%

\begin{document}

\title[Article Title]{Multi-Variable Batch Bayesian Optimization in Materials Research: Synthetic Data Analysis of Noise Sensitivity and Problem Landscape Effects}


\author[1]{\fnm{Imon} \sur{Mia}}\email{imon.mia@utdallas.edu}

\author[2,3]{\fnm{Armi} \sur{Tiihonen}}\email{armi.tiihonen@gmail.com}

\author[1]{\fnm{Anna} \sur{Ernst}}\email{anna.ernst@utdallas.edu}

\author[1]{\fnm{Anusha} \sur{Srivastava}}\email{anusha.srivastava@utdallas.edu}

\author[3]{\fnm{Tonio} \sur{Buonassisi}}\email{buonassi@mit.edu}

\author*[1]{\fnm{William} \sur{Vandenberghe}}\email{william.vandenberghe@utdallas.edu}

\author*[1]{\fnm{and Julia W.P.} \sur{Hsu}}\email{jwhsu@utdallas.edu}

\affil*[1]{\orgdiv{Department of Materials Science and Engineering}, \orgname{University of Texas at Dallas}, \orgaddress{\city{Richardson}, \postcode{75080}, \state{TX}, \country{USA}}}

\affil[2]{\orgdiv{Department of Applied Physics}, \orgname{Aalto University}, \orgaddress{\city{Espoo}, \country{Finland}}}

\affil[3]{\orgdiv{Department of Mechanical Engineering}, \orgname{Massachusetts Institute of Technology}, \orgaddress{\city{Cambridge}, \state{MA}, \country{USA}}}


\abstract{Bayesian Optimization (BO) is increasingly used to guide optimization tasks in materials science research. To emulate real-world experiments, we perform batch BO simulation of two test functions with six design variables and varying noise levels. A needle-in-a-haystack landscape (Ackley function) characterizes searching for a desired but unusual materials property, and a landscape with nearly degenerate optima (Hartmann function) is typically of materials fabrication process. We test different acquisition functions, batch picking methods, and exploration hyperparameter values, and present learning curves for various performance metrics and visualizations to effectively track the optimization progress. Optimization outcomes are shown to depend on noise and the search landscape. Conducting synthetic studies enables researchers to estimate experimental budgets before transitioning to the inherent uncertainties of real experiments. The results and methodology of this study facilitate a greater utilization of BO in guiding experimental materials research, specifically in optimization problems with multiple design variables.}


\maketitle

\section{Introduction}\label{sec1}

Throughout history, breakthroughs in materials science and engineering have relied on the optimization of a rough fabrication process toward a fine-tuned set of process parameters. We can think about the shaping of tools during the stone age, the choice of the copper-tin composition in the bronze age, and the carbon content, furnace temperatures, and quenching conditions in the iron age. As our engineering capabilities advanced throughout the industrial and modern era, the number of design parameters that can be changed during material processing has increased. Identifying optimal values in the input variables will enable the discovery and production of new materials with the desired properties to be faster. Thus, accelerated optimization will enable rapid advancements in materials research.

Machine learning provides powerful new tools for optimization tasks. Bayesian optimization (BO) \cite{DBLP:journals/corr/abs-1012-2599,7352306,snoek2012practicalbayesianoptimizationmachine} has recently emerged as the leading method for efficient sequential optimization of black-box functions that are costly to evaluate, situations often encountered in materials science research. \cite{Liang2021,jin2024bayesian, Gongora} Experiments related to materials development --- both manual ones and those including varying levels of automation --- are typical examples of such costly optimization tasks.

Noise is inevitable in experiments and may have a substantial impact to the outcomes in high-dimensional optimization tasks.\cite{noack2020autonomousmaterialsdiscoverydriven,D1ME00154J,daulton2022,wang2022recentadvancesbayesianoptimization,Bellamy} Unfortunately, most of the popular BO algorithm choices implemented in open BO code repositories have been developed and benchmarked in scenarios with negligible noise. Therefore, it is unclear if the same choices of algorithm, acquisition function, or hyperparameter perform well in noisy, high-dimensional experimental tasks encountered in materials science or how the results vary for different problems. A systematic evaluation of the effects of noise on the outcome is thus needed and investigated in this work. We also discuss how to simulate noise to reflect the general signal-to-noise level in the experiments, as this is relevant for planning BO campaigns in the laboratory. 


In this work, we develop a framework to elucidate every step in Bayesian optimization (BO), enabling its use as an evaluation tool in simulations and as a debugging tool for materials science experiments. We select two six-dimensional (6D) test functions to represent commonly encountered optimization problems in materials science: the Ackley function (a needle-in-a-haystack landscape) and the Hartmann function (which has a local minimum close in value to the global minimum). We examine how increasing noise levels systematically affects optimization outcomes across these two contrasting landscapes. Our benchmarking studies investigate how the choice of optimization metrics, acquisition functions, hyperparameters, and batch-selection methods affect BO outcomes. We also present methodology to visualize the optimization progress in such high-dimensional problems, putting both inputs and objective at equal footing.

Testing BO in simulation with synthetic data before experiments helps researchers learn the method, troubleshoot issues, and verify their experimental budget is adequate. By systematically examining each component of the BO framework, this paper addresses key barriers that materials scientists face when implementing BO. This work bridges the gap between theoretical machine learning literature and practical applications by demonstrating how to build robust, high-dimensional, and noise-aware BO workflows for real experiments, enabling materials scientists to confidently adopt BO as a practical tool for their research.


\section{Bayesian Optimization Background}

In BO, data are collected iteratively with model guidance. The aim is to maximize the values of an objective function (e.g., mechanical strength of additively manufactured samples \cite{Gongora}) by determining the optimal values for each design variable (such as the angle and diameter of support bars within the structure \cite{Gongora}). There are two key parts in the BO method: a surrogate model and an acquisition function. A surrogate model is rebuilt using the accumulated dataset at each iteration of optimization. The next point(s) to acquire new data are selected from the domain consisting of the design variables based on the surrogate model according to a user-chosen acquisition policy. Specifically, the selection of the next point is performed through evaluating the acquisition function. The choice of acquisition function and its exploration hyperparameter determines the balance between the exploitation of areas that have previously performed well and the exploration of the unknown regions. While the concept of BO is simple, {\it i.e.,} to find the input values (aka design variables) that maximizes the unknown black-box function, many subtleties of BO can befuddle new users. 

It is important to choose a function that represents the search landscape of the research problem. Many materials optimization problems have the needle-in-the-haystack landscape which we emulate here with the Ackley function.\cite{Siemenn2023} Some examples of needle-in-the-haystack landscapes include searching for auxetic materials with a negative Poisson's ratio,\cite{deJong2015} materials with high thermoelectric figure of meric (ZT),\cite{Hinterleitner} and high-entropy alloys that exhibit both high strength and high ductility.\cite{Li2016} The common thread is the search for a highly unusual also highly desired materials properties, which is not common in nature. The Ackley function has its optimal value centered at the origin with a rapid drop-off to fluctuating background values that occupy most of the search space, which represent that the optimal objective value is highly sensitive to small changes in input parameters.

In contrast, false optimal problems are more commonly encountered in process optimization problems,\cite{Siemenn2023} which we emulate using the Hartmann function. In such problems, the optimization space may contain multiple competing high-performing configurations (local optima) alongside a global optimum, posing challenges for efficient search. Examples of this type of  landscapes include optimizing deposition parameters for perovskite solar cells,\cite{Liu2022} searching for stable perovskite composition,\cite{Sun2021}, synthesizing silver nanoparticle by microfluidics,\cite{mekki-berrada2021two} and identifying print parameters for enhanced output quality.\cite{deneault2021toward} These types of problems are well represented by the Hartmann function, characterized by gradual variations but containing deceptive local optima that may mislead optimization algorithms.

While many materials optimization problems tend to have a needle-in-a-haystack landscape and many process optimization problems exhibit a false optimal landscape,\cite{Siemenn2023} the division is not strict. For example, in a case of molecular discovery and drug design, Aldeghi showed that the structure-properties relationship landscapes are dramatically different for half-life, hydration free energy, and hepatocyte clearance.\cite{Aldeghi2022} In high entropy materials, the properties that follow the rule of mixtures such as bulk modulus and thermal expansion coefficient would likely have a Hartmann-type landscape, while those that depend on specific microstructures, such as high thermal conductivity, would be likely searching for a needle in a haystack.\cite{Toher2023} Thus, the choice of which function to adopt as the surrogate model for the experiment will rely on the domain expertise of the researchers. Nevertheless, we believe that Ackley and Hartmann functions cover most materials science problems. Therefore, we systematically investigate how BO performs within Ackley and Hartmann landscape types.

BO operation is highly affected by its selected hyperparameters. Gaussian process regression (GPR) \cite{Rasmussen2006Gaussian, gpy2014} machine learning model is the most popular BO surrogate model because it is a probabilistic model that outputs both the predicted value of the objective function (called in this context as posterior mean) and the uncertainty of the prediction (called in this context as posterior uncertainty). Thus, it is straightforward to exploit areas where predicted values are promising and explore areas where uncertainty is high. The choice of the GPR kernel and its hyperparameters --- amplitude and lengthscales --- represents the user’s beliefs on what the objective function looks like. Additionally, the GPR noise variance setting should reflect the noise level in the experiment. However, GPR is frequently used for materials science optimization problems so novel that the properties of the underlying objective function --- e.g., its range and general shape, the number of optima, or the noise level --- are not known beforehand.

The choice of acquisition function presents another confusion to new users due to wide selection in literature and BO code packages. Expected improvement (EI)\cite{Janusevskis2012,ament2024} and upper confidence bound (UCB) \cite{271617,DBLP:journals/corr/abs-0912-3995} are two commonly used ones. Most BO literature --- especially those that are mathematical and theoretical --- uses EI. In contrast, UCB has been shown to be effective in experimental problems.\cite{SUN20191437,XU2023112055} Users are challenged also by how to set the exploration hyperparameter value, which controls the tradeoff between exploitation and exploration in the acquisition function.

Additionally, practical material science optimization tasks often depend on many materials components and processing conditions, {\it i.e.}, they are multi-variable problems, which is called high-dimensional problems in machine learning. Since the input parameters may not be independent of each other, it is pertinent to optimize them simultaneously to understand the interactions and dependencies better.\cite{Cao2018} With the increasing adaptation of automated/autonomous experiments, the correct implementation of high-dimensional BO algorithms becomes especially critical because human supervision is fragmentary in these cases. 

Due to experimental complexity and sampling limitations, most non-automated experiments, still the most widely used method, involve three to six input parameters. This is reflected in the experimental BO literature to date which has been limited to low-dimensional problems. For instance, Gongora et al. applied BO in a 4D space to optimize the toughness of mechanical structures by varying the number of columns, the radius and wall thickness of a column, and the twist angle.\cite{Gongora} Xu et al. optimized flexible solar cell efficiency by tuning 2 precursor variables (concentration and additive volume) and 2 processing variables (photonic curing intensity and length the light is on).\cite{XU2023112055} Similarly Xie et al. tuned two materials variables (molar ratio and volume) and two process variables (DC voltage and reaction duration) for metal–organic framework synthesis.\cite{Xie_Yunchao_Zhang}  Deneault et al. used BO to optimize four syringe extrusion parameters (prime delay, print speed, x- and y-position) for additive manufacturing.\cite{deneault2021toward}  Mekki-Berrada et al. explored five microfluidic parameters for silver nanoparticle synthesis.\cite{mekki-berrada2021two} In this study, we systematically investigate BO performance in six dimensions (6D) using Ackley and Hartmann test functions as benchmarking for materials science optimization problems.

While 6D is already beyond many experimental BO reports, the number of input variables may span to 8–15 in autonomous labs. Scaling to such regimes introduces new challenges, including kernel decay and degraded uncertainty quantification in GPR models. However, dimensionality reduction through feature selection or extraction can lower the number of important features to a practical level. These can be implemented using ARD kernels \cite{Rasmussen2006Gaussian} to suppress irrelevant inputs or scalable surrogates such as random forests\cite{hutter2011sequential}. These provide a pathway to adapt our framework for more complex, higher-dimensional materials systems.

Two large differences exist between experimental and synthetic data optimizations that are relevant for designing experimental BO: (1) experiments are often performed in batches, {\it i.e.}, processing multiple samples at once, to save materials cost or time; and (2) real-world experimental data contain noise. Most BO work in general machine learning literature are performed for sequential optimization, i.e., picking a single sampling point for the next experiment based on the highest acquisition function value at each iteration of BO. When performing batch BO, the remaining points after the first pick should be selected to be meaningfully different from the already-sampled points; this is not straightforward because we do not gain any new information between selecting the first point and the rest of the batch. Generally there are two approaches for batch picking: serial and parallel.\cite{Mia2025} In the serial approach, the first pick is performed by choosing the maximal acquisition function value, the same as non-batch BO, and selecting a strategy, categorized as penalizing, exploratory, or stochastic,\cite{gonzalej,Ginsbourger2010,Ginsbpurger1,GULER2021827} to pick the rest of the batch. More advanced true parallel algorithms  integrate over a joint probability density function using a generalized non-batch acquisition function.\cite{Balandat2020} In this work, we focus on the serial approach since it is more intuitive and straightforward and does not require advanced machine learning knowledge. We examine three batch-picking methods: Local Penalization (LP),\cite{gonzalej} Kriging Believer (KB),\cite{Ginsbourger2010} and Constant Liar (CL),\cite{Ginsbpurger1,GULER2021827} representing penalizing, exploratory, stochastic strategy, respectively, and commonly implemented in open BO packages.\cite{hunt2020batch} Yet, their performance across different landscape, acquisition function, and noise level is not clear. 

There are many kinds of noise in real-world problems, but in general, they can be characterized as random fluctuations and time-dependent variance. The former affects the precision of the measurements and is the predominant variation for samples made in one single batch, {\it i.e.}, at the same time. The latter is often seen in samples made in different batches, which may be made at different times, by different researchers, or caused by tool degradation, and determines the accuracy of the result. In this work, we will only focus on random noise because they are present in all experimental data. 

Because of the high dimensionality, the optimization progress is challenging to visualize; hence, assessing the progress of the algorithm-guided experiments is not straightforward. A widely used approach in the literature is to monitor the surrogate model's optimum objective value . However, this single metric for the BO is not capable of describing the results of the optimization campaign in detail. Furthermore, this approach focuses on the output ($y$), while in experiments, the researchers can only controll the design variable input values (${\bf X}$). Implicitly there is an assumption that convergence of the output implies convergence of the input, an assumption not verified until the suggested point has actually been characterized. In this work, we track the optimization progress for both ${\bf X}$ and $y$ and calculate their instantaneous and cumulative regrets. 

\section{Method}\label{sec2}

\subsection{Synthetic Objective Functions for Benchmarking and Analysis of Batch Bayesian Optimization}

\begin{figure*}[!h]
    \centering
    \includegraphics[width=0.7\linewidth,keepaspectratio]{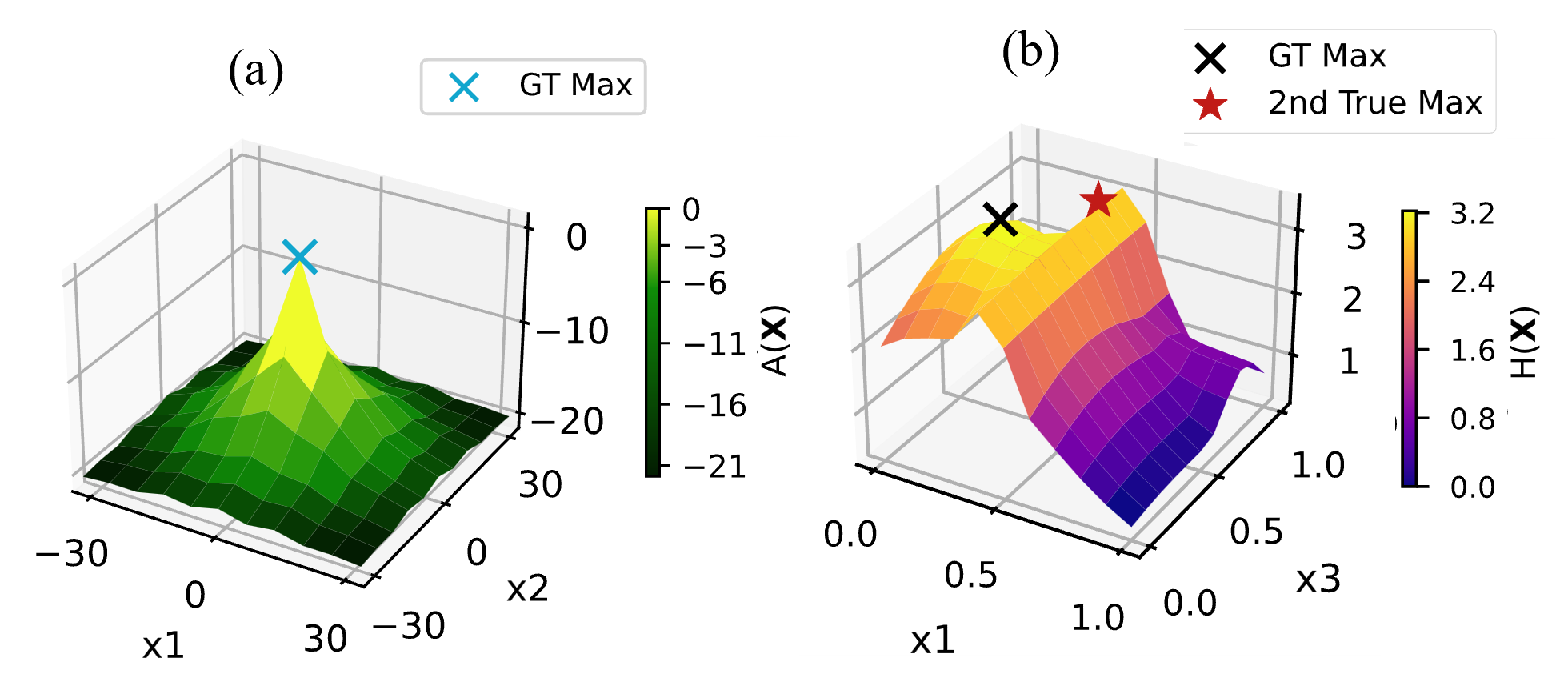}
    \caption{: Visualization of the 3D representation of the ground truth of (a) Ackley function, where x1,x2 variables are projected in 3D representation, and of (b) Hartmann function, where x1,x3 variables are projected in 3D representation. All the other input variables are chosen so that the maximum A({\bf X}) for a given (x1, x2) is shown for Ackley and the maximum H({\bf X}) for a given (x1, x3) is shown for Hartmann. The global maximum is labeled as the ‘GT max’ (cyan cross for Ackley, black cross for Hartmann) both functions and Hartmann function 2nd maxima labeled as the '2nd True Max' (red star).}
    \label{fig:1}
\end{figure*}

To implement test scenarios relevant in materials science, we consider two different types of synthetic test functions:
(1) an Ackley function\cite{simulationlib}
\begin{multline}
A({\bf X}) = 20 \left(\exp\left(-0.2 \sqrt{\frac{1}{d} \sum_{i=1}^{d} x_i^2}\right)-1\right) + \exp\left(\frac{1}{d} \sum_{i=1}^{d} \cos(2\pi x_i)\right) - \exp(1)
\label{1}
\end{multline}
where the dimension $d$ = 6, the range of $A({\bf X})$ is $[-22.3, 0]$ whereas the domain ${\bf X}$ is a 6D hypercube spanning $[-32.768, 32.768]$ along each dimension. Note that compared to Ref.~\cite{simulationlib}, we changed the sign of the Ackley function so that the optimization problem is a maximization problem. The Ackley function, illustrated in Fig.\ref{fig:1}(a)(only x1,x2 pair variables projected on Ackley function GT) and Fig.~\textbf{S1}(all pairs of variables projected  on Ackley function GT), has one sharp global maximum $A({\bf X}_{\rm max})=A({\bf 0})=0$ at the origin. 
The Ackley function thus represents a needle-in-the-haystack-type problem. Ackley-type functions are sometimes called heterogeneous functions,\cite{diouane2023trego,eriksson2020scalableglobaloptimizationlocal} where the convex region around the maximum occupies a minuscule fraction of the parameter space. Specifically, greater than 99.99\% of the hypervolume of the domain has $f({\bf X})<-15$, or, equivalently and more precisely, when randomly sampling ${\bf X}$, the probability of finding $f({\bf X})>-15$ is less than $4\times10^{-5}$. Some examples of such problem landscapes include searching for auxetic materials with a negative Poisson's ratio,\cite{deJong2015} materials with high thermoelectric figure of meric (ZT),\cite{Hinterleitner} and high-entropy alloys that exhibit both high strength and high ductility.\cite{Li2016} The common thread is the search for a highly unusual also highly desired materials properties, which is not common in nature. 

(2) a Hartmann function\cite{simulationlib}
\begin{equation}
H({\bf X}) =  \frac{1}{1.94} \left[ 2.58 + \sum_{i=1}^{4} \alpha_i \exp\left(-\sum_{j=1}^{d} A_{ij} ( x_j - P_{ij})^2 \right)\right]
\label{2}
\end{equation}
where $d$ = 6, $\alpha = (1.0,1.2,3.0,3.2)^T$ , $A_{ij}$ and $P_{ij}$ are $4\times6$ matrices defined in Supplementary section 1. The range of $H({\bf X})$ is $[0, 3.32237]$ whereas the domain ${\bf X}$ is a 6D unit hypercube spanning from $[0,1]$ along each dimension. We visualize the Hartmann function in Fig.\ref{fig:1}(b)(only x1,x3 variables pair projected on Hartamnn function GT) and Fig.~\textbf{S2} ( all pairs of variables projected on Hartmann function GT). The Hartmann function has one global maximum \( H({\bf X}_{\max}) = 3.32237 \),  
located at \( {\bf X}_{\max} = (0.20169, 0.150011, 0.476874, \)  
\( 0.275332, 0.311652, 0.6573) \). A second local maximum  
\( H({\bf X}_{\max,2}) = 3.20452 \) is found at  
\( {\bf X}_{\max,2} = (0.20169, 0.150011, 0.476874, \)  
\( 0.275332, 0.311652, 0.6573) \). The gradients near the maximum of the Hartmann function change much more gradually compared to the Ackley function, and we say that the Hartmann function is non-heterogeneous. The Hartmann function represents a shallow maximum and a landscape where optimization can easily get trapped in a local maximum far away from the global maximum. Example optimization problems of this type of search landscape include optimizing deposition parameters for perovskite solar cells,\cite{Liu2022} searching for stable perovskite composition,\cite{Sun2021}, synthesizing silver nanoparticle by microfluidics,\cite{mekki-berrada2021two} and identifying print parameters for enhanced output quality.\cite{deneault2021toward} ]

\subsection{Main Parameters of Bayesian Optimization to Be Considered}
The most utilized and investigated surrogate model for BO is GPR. The posterior distribution of a GPR model at a point ${\bf X}$ is  $f({\bf X}) \sim N({\bf m}, {\bf K})$ where $N$ refers to the normal distribution, ${\bf m}$ the mean vector, and ${\bf K}$  the covariance matrix. Given a set of observed data points  ${\bf D}^n = \left\{{\bf X}^i, y^i\right\}_{i=1}^{n}$, $i$ being the sample index and $n$ being the number of samples that has been evaluated so far, for a new point ${\bf X}^{i+1}$, the posterior predictive distribution can be computed as,  $y^{n+1} \sim N(\mu_{\bf D}({\bf X}^{n+1}), (\sigma_{\bf D})^2({\bf X}^{n+1}))$, where $\mu_{\bf D}$ is the predicted mean and $(\sigma_{\bf D})^2$ is the variance, computed from ${\bf D}^n$. 

The utility function $u({\bf D})$ quantifies the quality of the the dataset ${\bf D}$ where a higher $u$ indicates that ${\bf D}$ is of higher quality and better able to specify where the global maximum is.\cite{garnett_bayesoptbook_2023,Lizotte} The most naive choice for the utility function is the best encountered data point ${\rm Max}(y)$; however, this utility function can yield high values due to noise which is not meaningful. Instead, we use the utility function
\begin{equation}
u({\bf D}) = {\rm Max}(\mu_{\bf D} ({\bf X}))
\label{4}
\end{equation}
where, $\mu_{\bf D}$ is the mean posterior computed using ${\bf D}$, and the maximum (${\rm Max}$) is considered over all ${\bf X}$ in ${\bf D}$. Eq.~\ref{4} approaches $u({\bf D}) = {\rm Max}(y)$ in the absence of noise, but will be a better utility function in the presence of noise.

We consider the EI and UCB acquisition policies. The EI acquisition function is expressed explicitly in terms of the predicted mean $\mu_{\bf D}$  and variance $(\sigma_{\bf D})^2$ as
\begin{equation}
{\rm EI}({\bf X}|\mu_{\bf D},(\sigma_{\bf D})^2) = (\mu_{\bf D}({\bf X})- u({\bf D}) - \xi)\phi(Z({\bf D})) + \sigma_{\bf D}({\bf X})\varphi(Z({\bf D}))
\label{5}
\end{equation}
\begin{equation}
Z({\bf D}) = \frac{\mu_{\bf D}({\bf X}) - u({\bf D}) - \xi}{\sigma_{\bf D}({\bf X})}
\label{6}
\end{equation}
where $\phi(Z)$ indicates cumulative distribution function (CDF) and $\varphi(Z)$ indicates probability distribution function (PDF) of the GPR surrogate model. $\xi>0$ is an exploration hyperparameter; the larger the $\xi$, the more aggressive the exploration. The UCB acquisition function is defined by,
\begin{equation}
\rm UCB({\bf X}|\mu_{\bf D},\sigma_{\bf D}) = \rm \mu_{\bf D}({\bf X})+ \beta \times \sigma_{\bf D}({\bf X}), \beta > 0
\label{7}
\end{equation}
where $\beta$ represents an exploration hyperparameter. In some literature and software packages, the second term of UCB is expressed as $\sqrt{\beta} \sigma$. Here we use the same convention as in Emukit with linear $\beta$. We selected UCB and EI because they are used widely, have been implemented in multiple BO code packages, and can tune exploration and exploitation easily via hyperparameters $\beta$ and $\xi$, respectively. In this work, we benchmark both acquisition functions with a range of exploration/exploitation hyperparameter values to evaluate how they affect BO learning outcome.

To mimic experiments in fabricating/testing of multiple samples in a round, we perform batch BO with a batch size of four. We compare three serial batch-picking methods: Local Penalization (LP),\cite{gonzalej} Kriging Believer (KB),\cite{Ginsbourger2010} and Constant Liar (CL),\cite{Ginsbpurger1,GULER2021827} representing penalizing, exploratory, stochastic  strategies and commonly implemented in open BO packages. \cite{hunt2020batch} Although BO was created as a sequential optimization tool, employing batch evaluation is critical to leverage parallelization. KB and CL are greedy versions of more sophisticated true parallel global optimization.\cite{Balandat2020} 


\subsection{Benchmark Approach }

\begin{figure*}
    \centering
    \includegraphics[width=\linewidth]{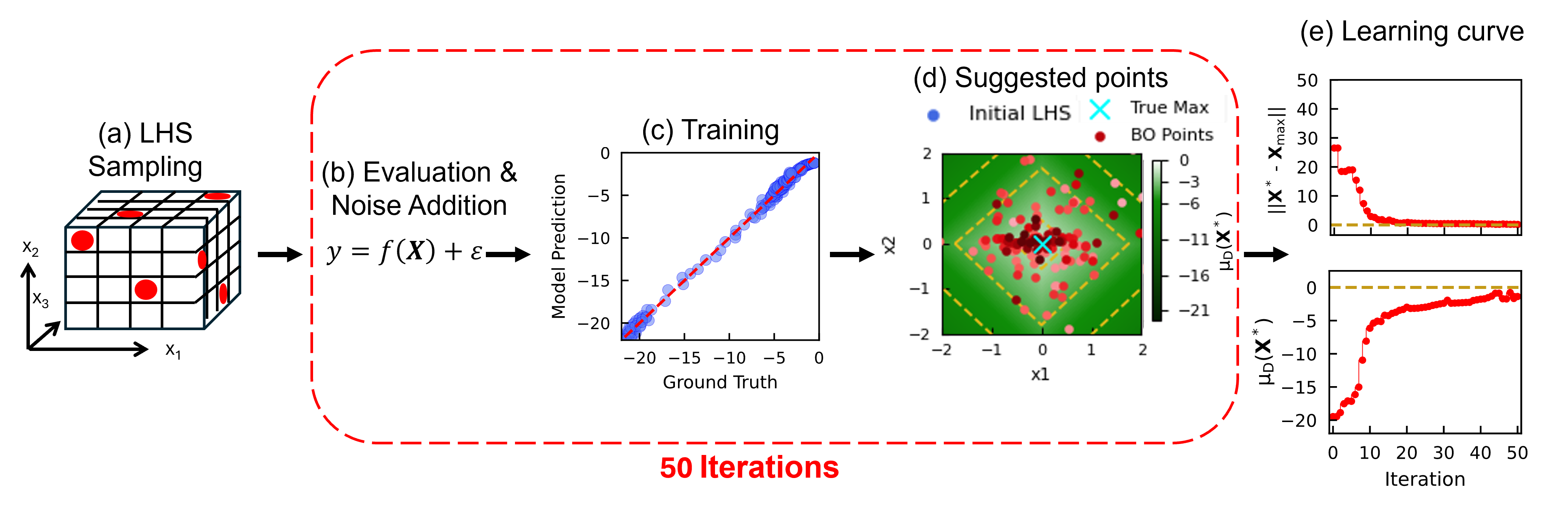}
    \caption{The workflow in the batch BO benchmarking: (a) LHS of the 6D input variable space to pick the starting points for the BO, (b) evaluations of the analytical test function at the selected points with an option to include noise, (c) the surrogate GPR model training at each iteration of the BO learning, (d) picking a batch of input points for the next iteration. The whole BO learning runs 50 iterations to generate the (e) {\bf X} (top) and y (bottom) learning curves of the benchmark criteria selected for this work, tracking the distance of the surrogate model optimum point to the true optimum location and the value of the surrogate model optimum, respectively. This whole process is repeated for 99 different LHS initial samplings to collect statisticsBO.}
    \label{fig:2}
\end{figure*}

We benchmark batch BO using Emukit open source BO code package \cite{emukit2019,emukit2023} with the two test objective functions (A($\bf X$) and H($\bf X$)), two acquisition functions (EI and UCB) with respective hyperparameters, as well as three analytical batch-picking methods (LP, KB, and CL), with different levels of noise according to the workflow illustrated in Fig. \ref{fig:2}. The workflow involves initial sampling (Fig. \ref{fig:2}(a)), repeating the BO iterations which include evaluating the objective values at each sampled point (Fig. \ref{fig:2}(b)), model training (Fig. \ref{fig:2}(c)), and suggesting the input points to evaluate for the next batch of points (Fig. \ref{fig:2}(d)), and evaluating the learning curves after the end of BO, in our case after 50 iterations (Fig. \ref{fig:2}(e)). BO settings, apart from the benchmark variables, are kept the same across the benchmark simulations, as detailed next.

As illustrated in Fig. \ref{fig:2}(a), the BO was initialized with 24 ${\bf X}$ points using the LHS method. Figs. \ref{fig:2}(b)-(d)  show how noise (if any) is added to the objective function values at the sampled points, the BO model is trained based on the accumulated data points, and a batch of ${\bf X}$ is chosen based on the acquisition function and batch-picking method. We considered BO in a 6D space with 4 points per batch and analyzed the progress through learning curves after 50 iterations as illustrated in Fig. \ref{fig:2}(e). Internally, both the input ${\bf X}$ and their corresponding $y$ values are normalized to the respective ranges of the ground-truth functions, i.e., the input and output variable values are within the unit hypercube. In our figures, we present results scaled back to their actual range. 

Noise is generated according to,
\begin{equation}
y^i = \rm f({\bf X}^\textit{i}) + \epsilon^\textit{i}, \epsilon^\textit{i} \sim \textit{N}(0, (\sigma^i)^2)
\label{8}
\end{equation}
We introduce noise in two distinct methods to analyze its impact on BO. Following the common practice in the literature \cite{D1ME00154J}, noise was added as a percentage of the maximum GT value,
\begin{equation}
\sigma^i = {\rm Max}(y_{GT}) \times \textrm{proportion\; of\; noise}
\label{9}
\end{equation}
In experiments, the GT is unknown, and thus the noise level is typically characterized as signal-to-noise ratio (SNR). To represent SNR in the synthetic data, we argue that the kernel amplitude in the noiseless case represents the general level of the signal better than its global maximum value, and hence noise should be added as a percentage of the noiseless kernel amplitude,
\begin{equation}
\sigma^i = \rm \textrm{(kernel\; amplitude)}_{\epsilon=0} \times {\rm proportion\; of\; 
noise}
\label{10}
\end{equation}
The results of these two ways to incorporate noise are compared.

Model training (Fig.~\ref{fig:2}(c)) involves training a GPR surrogate model at each iteration of BO. In each scenario, the ARDMatern52 is used as the kernel function. Automatic relevance determination (ARD) kernels assume a different length scale for each dimension and tune them separately. \cite{frazier2018tutorialbayesianoptimization,neal2012bayesian} In a previous benchmark by Liang et al., the ARD kernels performed better than the non-ARD ones.\cite{Liang2021} The ARD Matern kernel function is:
\begin{equation}
\rm K_M(x1,x2) = \sigma^2 \frac{2^{1-\nu}}{\tau(\nu)} \left(\frac{\sqrt{2\nu} |x1-x2|}{L}\right)^\nu K_\nu \left(\frac{\sqrt{2\nu} |x1-x2|}{L}\right)
\label{11}
\end{equation}
where $\sigma^2$ is the amplitude parameter, L is a length scale parameter, $\tau(.)$ is the gamma function, $\rm K_\nu$ is a modified Bessel function of the second kind and $\nu$=5/2 refers to the smoothness of the function (lower value means more smooth).\cite{snoek2012practicalbayesianoptimizationmachine,li2024studybayesianneuralnetwork}
The kernel amplitude and length scales are hyperparameters that must be correctly tuned to construct a realistic GPR surrogate model. Kernel amplitude represents the value range in the surrogate model predictions, and length scale represents the correlation between two points in the input space $\bf X$ of the surrogate model. Additionally, GPR model has a Gaussian noise variance (GNV) hyperparameter, representing the uncertainty associated with each observation and, when applied to experiments, reflecting the noise level in the data. Therefore, the GPR surrogate model may overly smooth the observed data if the noise variance is too high, whereas the model may overfit the noisy observations if the noise variance is too low. Both cases lead to suboptimal decisions. All three hyperparameters are autotuned in the simulation and tracked with iterations.

The next four points to sample are suggested according to the acquisition function and batch-picking method (Fig. \ref{fig:2}(d)). We test LP, KB, and CL as batch-picking methods for both Ackley and Hartmann functions with both EI and UCB as the acquisition functions with optimized hyperparameters based on performance metrics. We find LP outperforms KB and CL for both functions independent of the choice of acquisition function. For details see Supporting Information appendix section D. For the rest of the paper, LP is used as the batch-picking method unless explicitly specified.

We always consider 99 different initial LHS samplings which we indicate with a subscript index $k$ and 50 iterations which we indicate with a superscript $i$. 
When an index is omitted, it implies the collection of all ${\bf X}$ under consideration.
We define ${{\bf X}_k^i}^*$ as the ${\bf X}$ associated with ${\rm Max}(\mu_{\rm D})$ up to the current iteration, the formal definition is
\begin{equation}
\mu_{\rm D}({{\bf X}_k^i}^*) = {\rm Max}_{j=1}^{24+4i}(\mu_{\rm D}({\bf X}_k^j))
\end{equation}
where the maximum is taken over all 24 evaluations from the LHS sampling and the $4i$ evaluations in the first $i$ iterations of the BO. $\mu_{\rm D}({{\bf X}}^*)$ is the same as ${\rm Max}(\mu_{\rm D}({\bf X})$ defined in Eq.~\ref{4}.

\subsection{Metrics for Performance Evaluation During Benchmarking }
To characterize the optimization of the BO results, we compute instantaneous regret (IR) for ${\bf X}$ and $y$: ${\rm IR}({\bf X}_{k}) = || {{\bf X}_k^{50}}^* - {\bf X}_{\rm max} ||$, and ${\rm IR}(y_k) = | \mu_{\rm D} ({{\bf X}_k^{50}}^*) - y_{\rm max}|$, and then average the IR over all 99 LHS samplings to establish statistical variation:
\begin{equation}
\langle {\rm IR}({\bf X}) \rangle = \frac{\sum_{k=1}^{99} || {{\bf X}_k^{50}}^* - {\bf X}_{\rm max}
||}{99}
\label{14}
\end{equation}
and
\begin{equation}
\langle {\rm IR}(y) \rangle = \frac{\sum_{k=1}^{99} | \mu_{\rm D} ({{\bf X}_k^{50}}^*) - y_{\rm max}|}{99}.
\label{15}
\end{equation}
To quantify the convergence rate, average cumulative regrets in X and y are calculated:
\begin{equation}
\langle{\rm CR}({\bf X})\rangle = \frac{\sum_{k=1}^{99} \sum_{i=1}^{50} || {{\bf X}_k^i}^* - {\bf X}_{\rm max}
||}{99}
\label{16}
\end{equation}
and
\begin{equation}
 \langle{\rm CR}(y)\rangle = \frac{\sum_{k=1}^{99} \sum_{i=1}^{50} | \mu_D ({{\bf X}^i_k}^*) - y_{\rm max}|}{99}.
\label{17}
\end{equation}
For all four metrics, smaller values indicate a better performance.

\section{Results and Discussion}\label{sec3}

To gain insight on the behavior of BO under conditions commonly met in experimental materials science applications, we repeated simulated benchmarks of BO under varying levels of noise for two analytical test objective functions illustrated in Fig. \ref{fig:1}: (a) A($\bf X$) representing a needle-in-a-haystack problem that could be encountered in, e.g., molecule property optimizations, and (b) H($\bf X$) representing a landscape with a false maximum that could be encountered in, e.g., material composition optimizations. BO was compared in these scenarios with two acquisition functions (EI and UCB) with respective hyperparameters, as well as three batch-picking methods (LP, KB, and CL). These options represent typical selections that need to be made while setting up an experimental materials optimization campaign. The benchmark workflow (Fig. \ref{fig:2}) can be used for any simulated test function that represents the researchers' beliefs of the experimental problem landscape based on the prior expertise and experiences of the researcher on the specific application, with noise levels that are estimated to be realistic for the experimental setting. Next, the observations from the benchmarks are detailed in noise-free and noisy scenarios for the two different experiment landscapes.

\subsection{Noise-free Scenario}
\subsubsection{Needle-in-a-Haystack Landscape, Ackley}

\begin{figure*}
    \centering
    \includegraphics[width=0.8\linewidth]{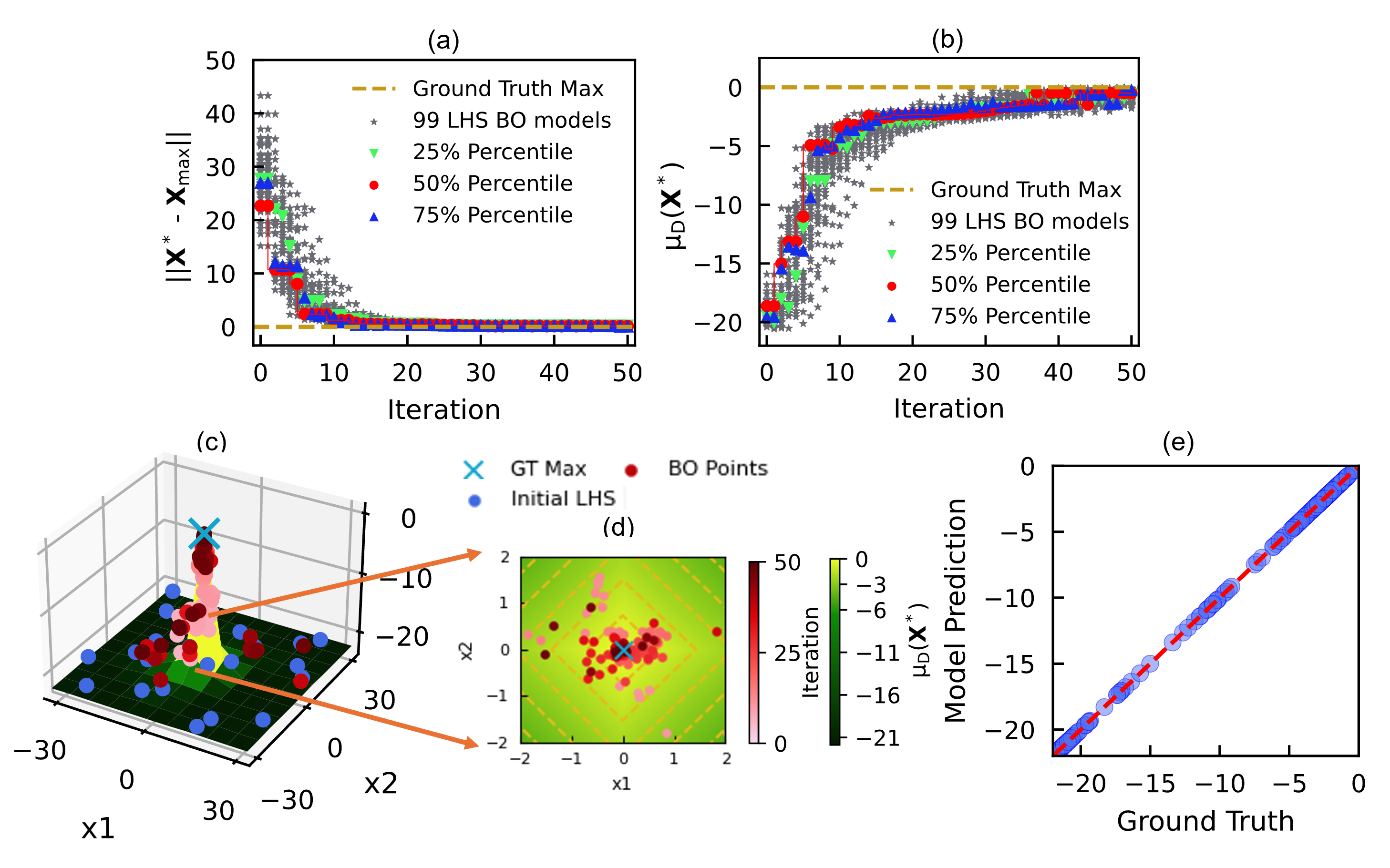}
    \caption{BO results using UCB with $\beta$ = 1 on noise-free Ackley function. Learning curve in (a) $||{\bf X}^* - {\bf X}_{\rm max} ||$ and (b) $\mu_{\rm D} ({\bf X}^*)$ for all 99 LHS initial starts. The 25th percentile (green triangle), 50th percentile (red circle) and 75th percentile (blue triangle) regions are highlighted to exemplify poor, median, and good LHS BO models, respectively. (c) Visualization of the 3D representation (x1 vs. x2) for the 50th percentile BO model, showing how BO iterations zero in on ${\bf X}_{\rm max}$ (cyan cross). Light blue circles are the 24 initial LHS selections and the pink to dark red points are training points (progressively darker). (d) Zoomed-in 2D heat map for variables x1 and x2 within the range [-2, 2] near ${\bf X}_{\rm max}$. (e) Parity plot of the 50th percentile BO model prediction vs. GT values for all 224 sampled points (LHS + 50 iteration at a batch size of 4).}
    \label{fig:3}
\end{figure*}

Noise-free optimizations are investigated first to determine the ideal performance of BO on our optimization tasks, and to illustrate the convergence of the BO in terms of our performance evaluation metrics. Fig. \ref{fig:3} shows the BO of the Ackley function using the UCB acquisition policy. Fig. \ref{fig:3}(a) shows $||{\bf X}_k^i - {\bf X}_{\rm max}||$, the deviation of the current optimal location from the GT maximum location (${\bf X}_{\rm max}$) as a function of iteration index $i$ for all 99 LHS samplings. Since the initial pick is random, none of the 99 samplings yield initial vectors close to ${\bf X}_{\rm max}$. Thus, $||{\bf X}^1-{\bf X}_{\rm max}||>15$ which is not surprising since the search space is a large hypercube measuring $[-32.768,32.768]^6$. After 10 iterations significant progress towards the optimum is made, $||{\bf X}^{10} - {\bf X}_{\rm max}||<10$ for all 99 samplings. At the last iteration ${\rm IR}({\bf X}_k)=||{\bf X}_k^{50} - {\bf X}_{\rm max}||<1$ and the estimated optimal inputs for all 99 samplings are very close to the GT maximum.

After 50 iterations, the 99 samplings are ranked based on ${\rm IR}({\bf X})$ from 1 to 99 percentile (worse to best). In Fig.~\ref{fig:3}, we highlight the evolution of the sampling that is ranked 25$^{\rm th}$ (poor outcome), 50$^{\rm th}$ (median outcome), and 75$^{\rm th}$ (good outcome) in green, red, and blue, respectively. The percentile ranking is made based solely on the result after the 50$^{\rm th}$ iteration, and therefore the order can change throughout the iterations. Indeed, upon careful inspection, we observe that the 75$^{\rm th}$ ranked sampling is further away from the GT compared to the 50$^{\rm th}$ ranked sampling for the first six iterations.

Fig. \ref{fig:3}(b) shows the $y$ value estimated by the surrogate GPR model, $\mu_{\rm D} ({\bf X}^*)$, as a function of BO iterations. At iteration 1, all $y$-value estimates are below $-15$, whereas around iteration 20 estimates have improved to be above $-5$, and at the final iteration the $y$-values are all close to 0, the maximum $A({\bf X})$ value.  Fig. \ref{fig:3}(c) shows a 3D representation of the surrogate model and scatter plot of the inputs of the 50$^{\rm th}$ percentile sampling on two of the six ${\bf X}$ dimensions. Along the input range from -32.768 to 32.768, the objective function ranges from - 22.3 to 0. Each dot indicates an input point; blue dots indicate initial LHS points and the red dots from light to dark represents the optimization progress. As iterations progress more points are sampled close to ${\bf X}_{\rm max}$, indicating that the BO exploits the optimum region. Fig.~\ref{fig:3}(d) shows the 2D heat map with focused domain range near maximum $A({\bf X})$ and reveals that most of the points for the Ackley function are sampled around ${\bf X}_{\rm max}$.

In addition to the analysis of performance metrics, the evolution of GPR hyperparameters is provided in Fig.~\textbf{S3}. We observe that all length scales converged to similar values as expected from the rotational symmetry around the origin of the Ackley function.  It should be noted that the exploration hyperparameter $\beta$ greatly affects BO convergence.\cite{garnett_bayesoptbook_2023,DBLP:journals/corr/abs-0912-3995}  Table ~\textbf{S1} shows the different metrics, $\rm \langle IR({\bf X})\rangle$, $\rm \langle CR({\bf X})\rangle$, $\rm \langle IR(y) \rangle$, and $\rm \langle CR(y)\rangle$, all normalized to the ${\bf X}, y$ ranges, for different $\beta$ values. We chose the $\beta$ that produces the smallest $\rm \langle IR(\bf X)\rangle$ which we determine as $\beta = 1$, which also yields the smallest $\langle {\rm CR}({\bf X})\rangle$ and $\langle {\rm CR}(y)\rangle$. 

Fig. \ref{fig:3}(e) shows the parity plot for the 50$^{\rm th}$ percentile sampling. The parity plot shows the GPR model posterior mean predictions as a function of their corresponding GT function values. The overall root mean square error (RMSE) of the posterior mean predictions on the training data is $\sim 4 \times 10^{-4}$. Normalizing the inputs to the unit hypercube and the outputs to $[0,1]$, the RMSE is $\sim 1.8 \times 10^{-5}$, reflecting the square root of the final value of the GNV hyperparameter of the model (Fig.~\textbf{S3}(h)). Although our data does not contain noise, a non-zero value is needed for the GNV hyperparameter to ensure convergence. We note that when the GNV hyperparameter is set to be proportional to the mean squared variance of the unnormalized $y$ values, the GPR model deviates from the GT function as if noise is added, indicating the GNV is too large. Fig.~\textbf{S4} shows how the learning outcomes are affected by the GNV values. Judiciously chosen GNV value ensures small RMSE value, i.e., the GPR model reproduces the GT Ackley function very well when data are without noise.
\begin{figure*}[h!]
    \centering
    \includegraphics[width=0.8\linewidth,keepaspectratio]{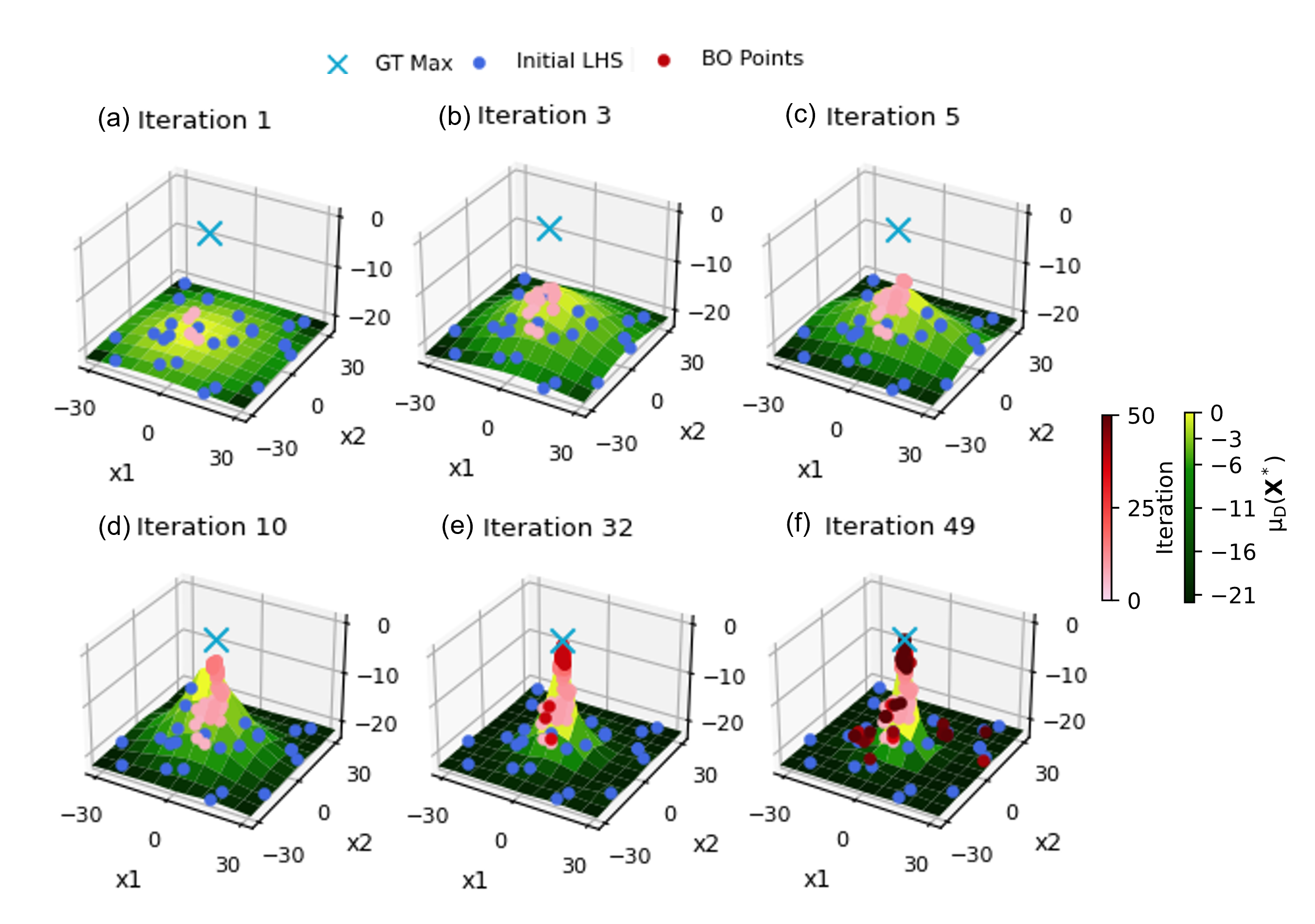}
    \caption{(a)-(f) 3D representations of Ackley test function (UCB with $\beta$=1) at different iterations. Blue points are the initial LHS points. Light pink points are sampled at earlier iterations and the dark red points are sampled at later iterations according to the color bar. }
    \label{fig:4}
\end{figure*}

To visualize the learning process even better, we provide Movie~(1) showing the evolution of the median GPR model where the iteration progress is represented through the time evolution. Fig. \ref{fig:4} shows still snapshots of the evaluated points and the associated GPR model taken from Movie~(1), clearly depicting how the model develops into something similar to the GT function as BO progresses and the number of data collected increases.

A concern with using the Ackley function for simulating BO is that its optimal point is located at the center of the domain, making it possible for the search to be accidentally expedited if using grid-sampling algorithm, due to the steepness of the maximum. Thus, we do not use the grid-sampling method in this work. In Fig.~\textbf{S5}, we test our BO algorithm on an Ackley function with a maximum off the center at (x1, x2, x3, x4, x5, x6). The convergence of BO is similar to Fig. \ref{fig:4} and the maximum is correctly identified.

\subsubsection{Landscape with Nearly Degenerate Maxima, Hartmann}

\begin{figure*}[h!]
    \centering
    \includegraphics[width=0.8\linewidth,keepaspectratio]{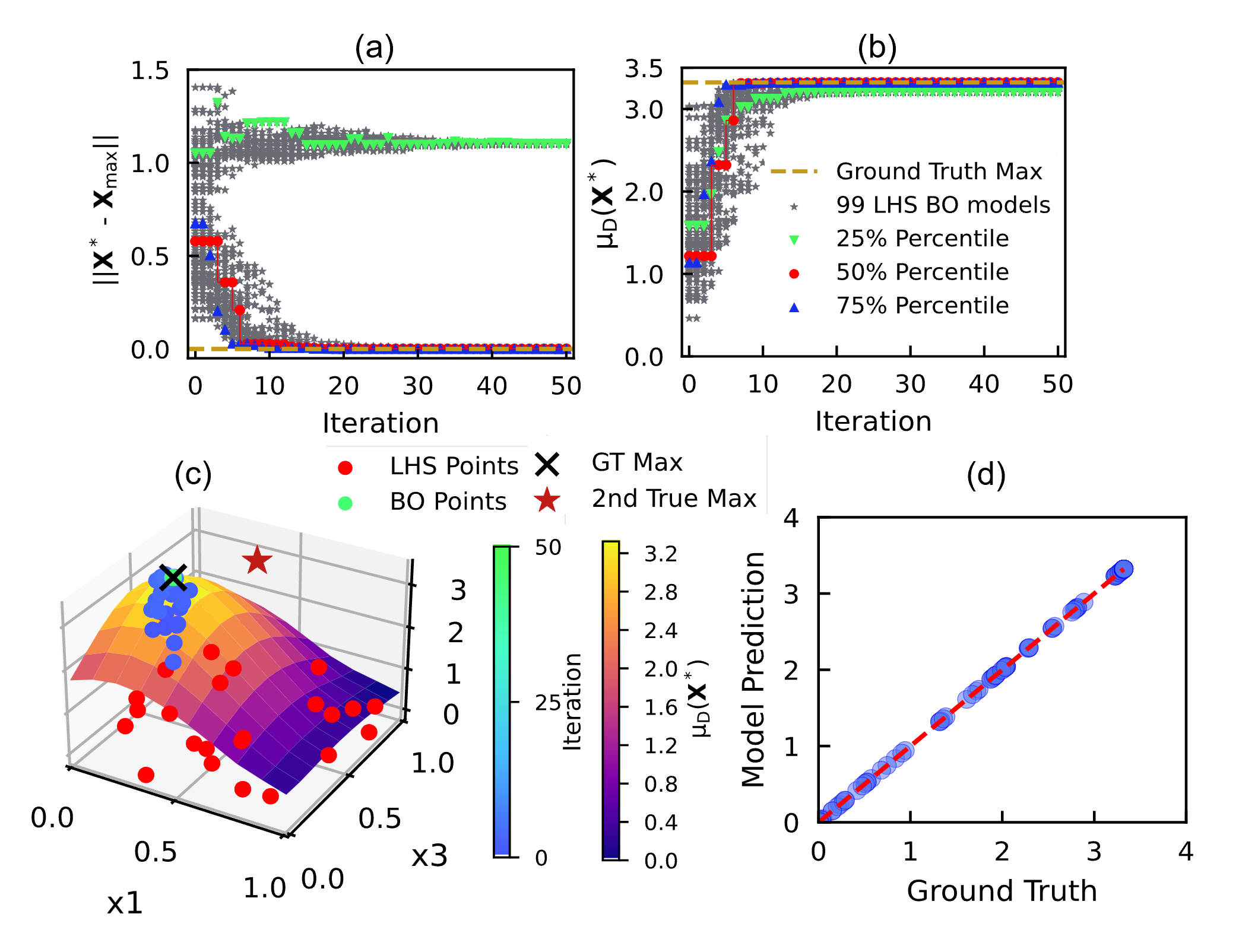}
    \caption{BO results using UCB with $\beta$ = 1 on noise-free Hartmann function. Learning curve in (a) $||{\bf X}^* - {\bf X}_{\rm max} ||$ and (b) $\mu_{\rm D} ({\bf X}^*)$ for 99 LHS BO models. The 25th percentile (green triangle), 50th percentile (red circle) and 75th percentile (blue triangle) regions are highlighted to exemplify poor, median, and good LHS BO models, respectively. (c) Visualization of the 3D representations for (x1 vs. x2) variables pair, for the 50th percentile BO model at the last iteration, showing how BO iterations zero in on ${\bf X}_{\rm max}$. Red circles are the 24 initial LHS selections and the blue to light green points are training points (blue progressively to green).(d) Parity plot of the 50th percentile BO model prediction vs. GT values for all 224 sampled points (LHS + 50 iteration at a batch size of 4).}
    \label{fig:5}
\end{figure*}

Fig.~\ref{fig:5}a shows $||{\bf X}_k^i-{\bf X}_{\rm max}||$ for the Hartmann test function using the UCB acquisition policy. Initially, the distance to the maximum $||{\bf X}^1_k-{\bf X}_{\rm max}||$ ranges from $0.2$ to $1.5$ in the unit hypercube. The distance to ${\bf X}_{\rm max}$ for the Hartmann function show relatively larger variance compared to the Ackley, which we attribute to those LHS sampling result in finding a optimum close to ${\bf X}_{\rm max,2}$ rather than ${\bf X}_{\rm max}$. In Fig.~\ref{fig:5}(a), for $\approx 75\%$ of the samplings, $||{\bf X}_k-{\bf X}_{\rm max}||$ converges to 0 while for the remaining $\approx 25\%$, $||{\bf X}_k-{\bf X}_{\rm max}||$ converges to $||{\bf X}_{\rm max,2}-{\bf X}_{\rm max}||\approx 1.1$. In some LHS samplings ${\bf X}_k$ is initially near ${\bf X}_{\rm max,2}$ but still ends up at ${\bf X}_{\rm max}$. Overall, because of the second maximum, it is significantly more challenging to find the GT max in the Hartmann landscape compared to the Ackley landscape.
Fig.~\ref{fig:5}b shows $\mu_{\rm D} ({\bf X}^*)$ as a function of iteration index for all 99 samplings. Initially, the values range from 0.5 to 3.0 (whereas the GT maximum is 3.32). The relative variance of $\mu_{\rm D} ({\bf X}^*)$ is much greater compared to the Ackley landscape. A number of LHS samplings find initial ${\bf X}$ which are already relatively close to $H({\bf X}_{\rm max})$. The reason is that the global maximum is wide and there is also a second maximum with almost as high function values, thus a much larger fraction of the domain yields values close to $H({\bf X}_{\rm max})$. After 10 iterations, all $\mu_{\rm D} ({\bf X}^*)$ are above 2.5 and then quickly converge to the GT maximum (3.32237) or the second ground truth maximum (3.20452).

The dependence of the BO results on the UCB exploration hyperparameter $\beta$ is shown in Table ~\textbf{S3}. Similarly to Ackley, $\beta$ = 1 produces the lowest $\rm \langle IR(y) \rangle$ when using UCB policy. Comparing Table ~\textbf{S1} and ~\textbf{S3} for $\beta$ = 1 shows that ${\rm \langle IR}({\bf X})\rangle$ and $\rm \langle IR(y) \rangle$ are  $\sim$ 210 times larger and $\sim$ 2 times smaller, respectively, for the Hartmann compared to the Ackley function. Our findings are in line with the benchmarking study of Liang et al.\cite{Liang2021}

In Fig.~\ref{fig:5}(c), the ${\bf X}$ evaluated at each iteration of the 50$^{\rm th}$ percentile (median outcome) LHS sampling are shown superimposed on the 3D representations of the GPR model. The red dots are LHS points and the blue-to-green dots represent BO optimization progress. The projections onto the other pairs of input variables are shown Fig. ~\textbf{S6}. The GPR model of the median sampling, hence all the percentile samplings above the median, accurately identify ${\bf X}_{\rm max}$ of the test function. The sampled points (blue to light green circles) converge on ${\bf X}_{\rm max}$ and no evaluated points appear near ${\bf X}_{\rm max,2}$ demonstrating that in this case, the BO model correctly identifies ${\bf X}_{\rm max}$.
The parity plot in Fig.~\ref{fig:5}(d) compares the median BO model predictions with the GT data for all the evaluated points. RMSE value for this plot is $2.6 \times 10^{-4})$. The normalized RMSE value is $7.8 \times 10^{-5}$, which corresponds well to the converged square root value of the GNV hyperparameter (Supplementary Fig. ~\textbf{S7}(h)).

Fig.~\textbf{S8} shows still snapshots  of the BO iterations for the 50$^{\rm th}$ percentile LHS sampling, projected onto the x1-x3 plane, whereas Movie~(2) shows the entire iterative progress of the BO. Note that in Fig.~\ref{fig:5}(c) the 3D representation is of the GP model of the final iteration whereas in Fig.~\textbf{S8}, the evolution of the GPR model is illustrated. Initially, the GPR model displays a very flat landscape without a peak and the $\mu_{\rm D} ({\bf X}^*)$ value is $<2$. As iterations progress, the GPR model develops a peak near the GT max ${\bf X}_{\rm max}$ (black cross). By the tenth iteration, the BO model correctly identifies the ${\bf X}_{\rm max}$. Note that there is no peak near the $2^{\rm nd}$ max ${\bf X}_{\rm max,2}$ (red star).

We also analyzed poor (25$^{\rm th}$ percentile) and good (75$^{\rm th}$ percentile) BO models based on the ranking of their final ${\rm IR}({\bf X})$ values (see Supplementary Movie~(3) and  (4), respectively). The poor GPR model has the optimum near ${\bf X}_{\rm max,2}$; on the contrary, the good model had the optimum exactly at the  ${\bf X}_{\rm max}$. The latter also had a comparatively faster convergence than the median LHS optimized model as indicated in Fig.~\ref{fig:5} (a) and (b).

\subsubsection{Comparison of Acquisition Function Performance}

The UCB and EI acquisition policies are compared for the Ackley and the Hartmann function optimization in Table~\ref{table:1} for three values of the exploration hyperparameters ($\beta$ for UCB and $\xi$ for EI, respectively) each. Comprehensive studies with all exploration hyperparameter values tested can be found in Table ~\textbf{S1} and ~\textbf{S3} for UCB and Table ~\textbf{S2} and ~\textbf{S4} for EI.
\begin{table}[h!]
\caption{Comparison between UCB and EI acquisition policies with noise-free Ackley and Hartmann test functions. Exploration parameter values ($\beta$ for UCB and $\xi$ for EI. respectively) are also listed.}
\label{tab:my-table}
\begin{tabular}{|c|c|c|c|c|c|c|c|}
\hline
\textbf{Test Function} &
  \textbf{\begin{tabular}[c]{@{}c@{}}Acquisition \\ Policy\end{tabular}} &  
  \text{\begin{tabular}[c]{@{}c@{}}$\rm \langle IR({\bf X}) \rangle$\\ 1 $\times 10^{-2}$\end{tabular}} &
  \text{\begin{tabular}[c]{@{}c@{}}$\rm \langle CR({\bf X})\rangle$\end{tabular}} &
  \text{\begin{tabular}[c]{@{}c@{}}$\rm \langle IR(y)\rangle$\\ 1 $\times 10^{-2}$\end{tabular}} &
  \text{\begin{tabular}[c]{@{}c@{}}$\rm \langle CR(y)\rangle$\end{tabular}} \\ \hline
Ackley                       & UCB ($\beta$ = 1) & 0.11 & 1.17 & 1.63  & 4.88\\ \hline
Ackley                       & UCB ($\beta$ = 15) & 12.4 & 10.1 & 33.8  & 21.5\\ \hline
Ackley                       & UCB ($\beta$ = 30) & 22.1 & 11.3 & 44.8  & 22.8\\ \hline
Ackley                      & EI ($\xi$ = 0) & 0.59 & 1.87 & 9.35 & 8.61\\ \hline
Ackley                      & EI ($\xi$ = 0.1) & 3.48 & 6.08 & 26.7 & 19.1\\ \hline
Ackley                      & EI ($\xi$ = 10) & 24.3 & 13.6 & 62.6 & 32.6\\ \hline
Hartmann                       & UCB ($\beta$ = 1) & 23.1 & 18.5 & 0.84 & 3.31\\ \hline
Hartmann                       & UCB ($\beta$ = 15) & 57.0 & 35.1 & 6.95 & 14.1\\ \hline
Hartmann                       & UCB ($\beta$ = 30) & 95.5 & 41.4 & 30.7 & 22.6\\ \hline
Hartmann                      & EI ($\xi$ = 0) & 33.0 & 18.9 & 0.91 & 3.10\\ \hline
Hartmann                      & EI ($\xi$ = 0.1) & 37.0 & 21.1 & 3.01 & 4.12\\ \hline
Hartmann                      & EI ($\xi$ = 10) & 66.0 & 33.9 & 28.9 & 19.3\\ \hline
\end{tabular}
\label{table:1}
\end{table}
Four BO performance evaluation metrics of averaged instantaneous and cumulative regrets (${\rm \langle IR}({\bf X})\rangle$, ${\rm \langle CR}({\bf X})\rangle$, ${\rm \langle IR}(y)\rangle$, and ${\rm \langle CR}(y)\rangle$) are compared: the lower the value, the better the performance. Regardless of the metric or objective function, UCB performs best when $\beta = 1$, and EI performs best for $\xi = 0$. For all except $\rm \langle CR(y)\rangle$ for Hartmann, UCB yields lower regret values than EI; in the case of Hartmann $\rm \langle CR(y)\rangle$, EI produces a slightly lower value (3.10) than UCB (3.31). Figs.~\textbf{S9} and ~\textbf{S10} clearly depict the superior performance of UCB compared to EI visually via the evolution of $||{\bf X}^*-{\bf X}_{\rm max}||$ and $\mu_{\rm D} ({\bf X}^*)$. The different is more significant for the Ackley function (Figs.~\textbf{S9}), with the largest difference observed in ${\rm \langle IR}({\bf X})\rangle$, where the UCB value is more than 5 times lower. Overall, we conclude that UCB significantly outperforms the EI, especially in terms of the convergence of $\mu_{\rm D} ({\bf X}^*)$, which is also in line with Liang et al.\cite{Liang2021}. 

To strengthen these descriptive comparisons, we examine the results at the end of optimization (50th iteration) across 99 LHS runs comparing using the best UCB ($\beta = 1$) and EI ($\xi = 0$). For the Ackley function, there is no overlaps between the UCB and EI results for either ${\rm IR}({\bf X})$ and ${\rm IR}(y)$, hence no need for further statistical tests. The median ${\rm IR}({\bf X})$ obtained by UCB and EI are $0.11 \times 10^{-2}$ and $0.60 \times 10^{-2}$, respectively, and the median ${\rm IR}(y)$ obtained by UCB and EI are  $1.4 \times 10^{-2}$ and $9.1 \times 10^{-2}$, respectively. For the Hartmann function, there are substantial overlaps between the results obtained by the two acquisition functions; hence, we  perform non-parametric Mann–Whitney U tests on them. The test reports a statistic ($U$) that measures how separated the ranks of the two groups are, and a probability value ($p$) that quantifies the likelihood of observing such a difference by chance.  For the Hartmann function, the median ${\rm IR}({\bf X})$ obtained by UCB and EI are  $2.5 \times 10^{-2}$ and $5.7 \times 10^{-2}$, respectively, yielding a $U = 2829$ and $p \approx 10^{-7}$; the median ${\rm IR}(y)$ obtained by UCB and EI are $0.22 \times 10^{-2}$ and $0.44 \times 10^{-2}$, respectively, yielding a $U = 4019$ and $p \approx 0.016$, still smaller than the $0.05$ used for statistically not able to reject the null hypothesis. Thus, for all cases, UCB and EI yield distinct performance distributions. The statistical tests further reinforce that the acquisition strategies differ meaningfully, with UCB outperforms EI.

\subsection{Noisy Scenario}

\subsubsection{Comparing Utility Functions in the Presence of Noise}

\begin{figure*}[h!]
    \centering
    \includegraphics[width=0.8\linewidth,keepaspectratio]{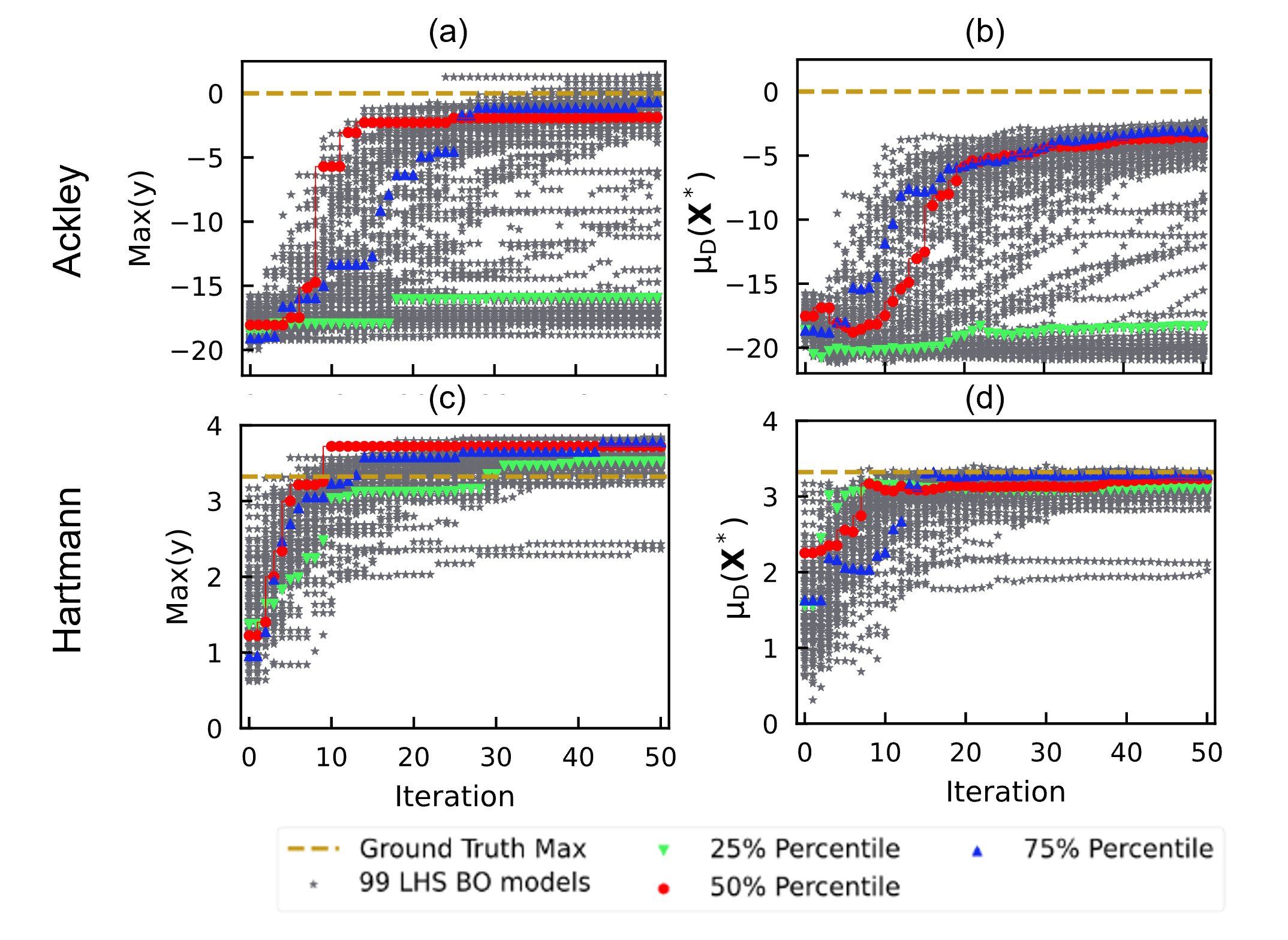}
    \caption{Learning curves of (a) Max(y) and (b) $\mu_{\rm D} ({\bf X}^*)$ for Ackley test function and of (c) Max(y), and (d) $\mu_{\rm D} ({\bf X}^*)$ for Hartmann test function using EI as acquisition function ($\beta$ =1) with 5\% noise level. The 25th percentile (green triangle), 50th percentile (red circle) and 75th percentile (blue triangle) BO models are highlighted to exemplify poor, median, and good LHS BO models, respectively. The yellow dash line represents the GT global maximum value of the test functions, $A({\bf X}_{\rm max})$ or $H({\bf X}_{\rm max})$.}
    \label{fig:6}
\end{figure*}

Fig.~\ref{fig:6}a-d compares the learning curves using ${\rm Max}(y)$ (a, c) vs $\mu_{\rm D} ({\bf X}^*)$ (b, d) utility functions with 5\% noise for both Ackley (a, b) and Hartmann (c, d) test functions. The slower convergence in Fig.~\ref{fig:6} compared to Figs.~\ref{fig:3} and \ref{fig:5} is due to the presence of noise in the data. The ${\rm Max}(y)$ results (left column) reveal that about 5 LHS samplings yield ${\rm Max}(y)>A({\bf X}_{\rm max})$ for Ackley, and the majority of LHS samplings produce ${\rm Max}(y)>H({\bf X}_{\rm max})$ for Hartmann. $y$ can exceed the GT maximum value because of the noise when  evaluating ${\bf X}$ close to the maximum or second maximum. In such a case, the learning curve is tracking the outliers, i.e., data point with the largest noise value, and the problem is exacerbated at higher noise levels. In contrast, when $\mu_{\rm D} ({\bf X}^*)$ is used as the utility function in the BO, almost no LHS sampling result in $\mu_{\rm D} ({\bf X}^*)$ exceeding the GT maximum value. Furthermore, the $\mu_{\rm D} ({\bf X}^*)$ learning curve of the Ackley test function (Fig.~\ref{fig:6}b) reveals more clearly that the algorithm has not yet converged to the optimum compared to the ${\rm Max}(y)$ curve (Fig.~\ref{fig:6}a). Overall, we conclude that using $\mu_{\rm D} ({\bf X}^*)$ as the utility function for the learning curves  is a more suitable measure of BO algorithm than Max(y), especially when noise is present.\cite{garnett_bayesoptbook_2023} We have tracked $\mu_{\rm D} ({\bf X}^*)$ in all other learning curves in the paper.

\subsubsection{Needle-in-a-Haystack Landscape, Noisy Ackley}

\begin{figure*}[h!]
    \centering
    \includegraphics[width=1\linewidth,keepaspectratio]{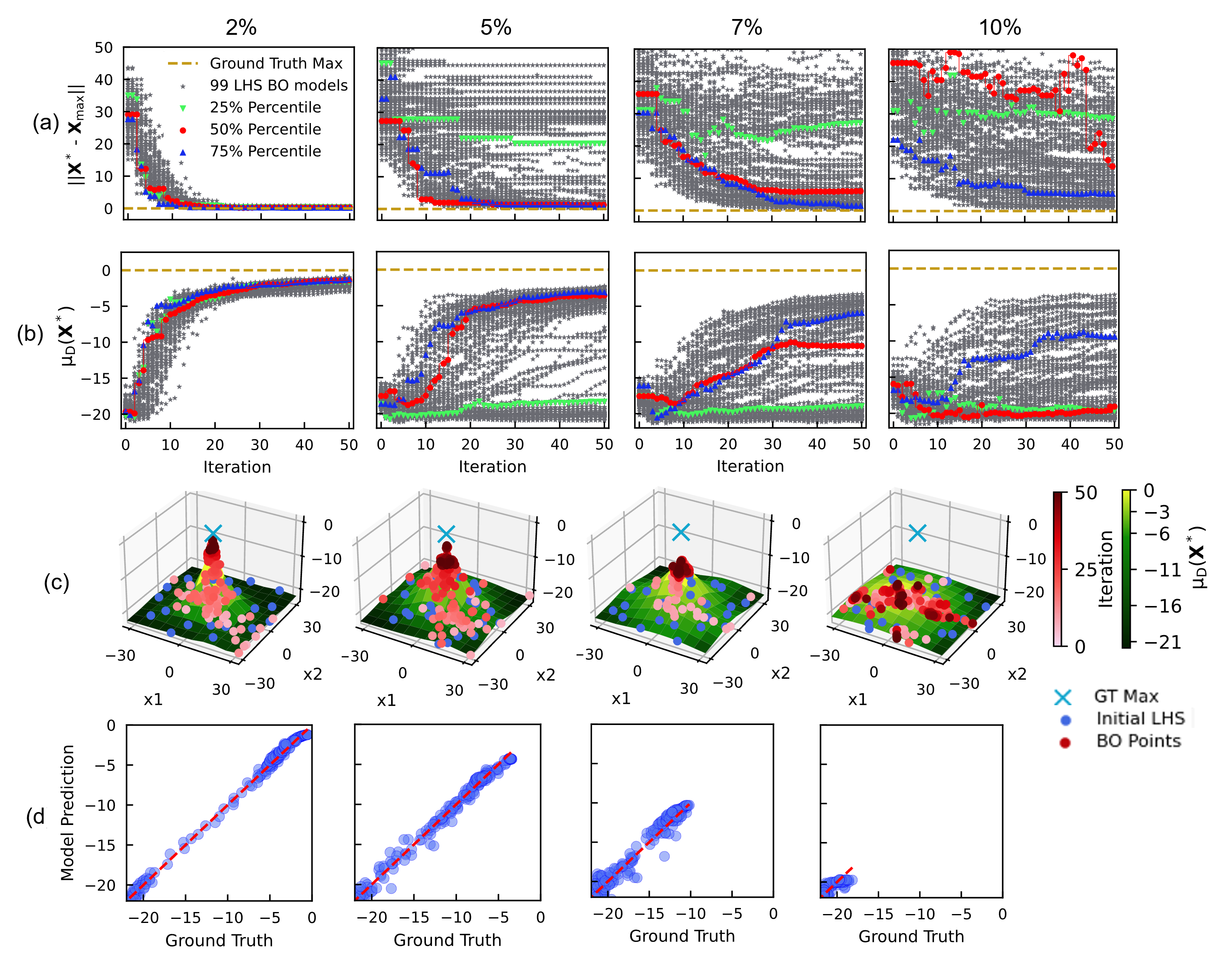}
    \caption{Learning curve of $||{\bf X}^* - {\bf X}_{\rm max} ||$ in top row (a) and $\mu_{\rm D} ({\bf X}^*)$ in the second row (b) for the Ackley function using LP as the batch-picking method and EI as the acquisition function for noise levels of 2\% ($\xi$ = 0), 5\% ($\xi$ = 0.05), 7\% ($\xi$ = 0.1), and 10\% ($\xi$ = 0.05) from left to right. For visualization, (c) 3D representations and (d) parity plots of the 50$^{\rm th}$ percentile BO models for the same noise levels. In the 3D representations, x1 and x2 are projected with 224 learning points, including 24 initial LHS points (blue) and 50 iterations (from light to dark red) of BO points.} 
    \label{fig:7}
\end{figure*}

Fig. \ref{fig:7} (a), and (b), show the learning curves of the Ackley test function for noise levels of 2\%, 5\%, 7\%, and 10\%, from left to right, using EI as the acquisition policy. With 2\% noise, the curves look similar compared to the no-noise case although at iteration 10, one LHS sampling does not satisfy $||{\bf X}^*-{\bf X}_{\rm max}||<10$ whereas all satisfied this criterion for the noiseless case. At 5\% noise, we observe that many samplings do not reach the GT maximum but the 50$^{\rm th}$ percentile sampling still reaches a value close to ${\bf X}_{\rm max}$. This indicates that a little more than half of the samplings find the global maximum location by the end of BO despite $\mu_{\rm D} ({\bf X}^*)$ is lower than $A({\bf X}_{\rm max})$. 

At 7\% noise, the $50^{\rm th}$ percentile sampling has $\mu_{\rm D}({\bf X}^*)\approx -10$ and even the $75^{\rm th}$ percentile sampling starts to deteriorate. Inspecting how far we are from the GT in ${\bf X}$, we observe a significant deviation with ${\rm \langle IR}({\bf X})\rangle \approx 5$ for the 50$^{\rm th}$ percentile, indicating that less than half the samplings get close to the ${\bf X}_{\rm max}$. Finally, for 10\% noise, even the $75^{\rm th}$ percentile does not find the maximum in ${\bf X}$ or $y$. Thus, noise has a clear impact on the convergence rate of BO in heterogeneous type functions, which is one of the key factors to consider when determining the experimental budget for optimization tasks.

Exploration hyperparameters are investigated for both UCB and EI acquisition functions as a function of noise level, based on the performance metric ${\rm \langle IR}({\bf X})\rangle$ (Table ~\textbf{S9} and ~\textbf{S10}). Comparing UCB $\beta=1$ and EI $\xi=0$, we find that for noise levels above 3\%, the EI acquisition performs better than UCB. The better performance using the EI acquisition function motivated us to compare BO results for different noise levels using EI in Fig.~\ref{fig:7}. Neither EI nor UCB have been initially designed to be used for noisy objectives,\cite{letham2018constrainedbayesianoptimizationnoisy} and ${\rm \langle IR}({\bf X})\rangle$ are indeed high  beyond 10\% of noise, indicating most of the LHS samplings have not resulted in a converged BO. Lately, acquisition functions have been designed specifically for noisy objectives, such as Noisy-EI,\cite{letham2018constrainedbayesianoptimizationnoisy,pmlr-v222-zhou24a} however, these have not yet been integrated into most BO packages widely used in materials science applications. Table ~\textbf{S11} shows the acquisition function and hyperparameter that yields the lowest ${\rm \langle IR}({\bf X})\rangle$ for each noise level, and reveals that for 2\% noise, UCB with $\beta=2$ is optimal whereas for 5\% and 10\%, EI with $\xi$ = 0.05 is optimal, and for 7\% noise, EI with $\xi=0$ is optimal. Generally, low $\xi$ values that lead to more exploitation perform clearly better in low-noise scenarios for Ackley, whereas in high-noise scenarios, the choice of $\xi$ does not affect the result drastically.  In the case of UCB, the effects of the exploration hyperparameter on the convergence are not as drastic but the low exploration hyperparameter values also perform better in low-noise scenarios. Thus, based on this benchmark, low exploration choices are a robust option for EI and UCB with Ackley-type objectives.

Fig.~\ref{fig:7}(c) shows the GPR model of the sampling with the $50^{\rm th}$ percentile ${\rm \langle IR}({\bf X})\rangle$ and Fig. \ref{fig:7}(d) shows the corresponding parity plots. For 2\% noise, the GPR model resembles the GT; for 5\%, the peak is slightly degraded; for 7\%, the peak is significantly degraded; and finally, for 10\%, the landscape is almost flat and the Ackley peak has all but disappeared. A zoom-in 2D heat map is shown in Fig.~\textbf{S15}, and  the GNV hyperparameters are shown in Fig.~\textbf{S16}. Supplementary Movie~(5) illustrates the progress of BO learning, showcasing the transformation from the initial to the final iteration of the $50^{\rm th}$ percentile sampling with 5\% noise. Overall, we observe good performance at 2\% noise with performance degraded as noise increases to almost no BO effectiveness at 10\% noise. It is clear from \ref{fig:7}(d) that the GPR model misses the peak when noise is above 5\%. For 10\% noise, the model was only covering $y$ values below -15, significantly lower than the peak. However, for Ackley, these regions constitute almost the entire ($\langle$ 99.99\%) GT hypervolume. This fact is important when we compare BO performance for Ackley vs. Hartmann in Fig. \ref{fig:12}.

\subsubsection{Landscape with Nearly Degenerate Maxima, Noisy Hartmann}

\begin{figure*}[h!]
    \centering
    \includegraphics[width=1\linewidth,keepaspectratio]{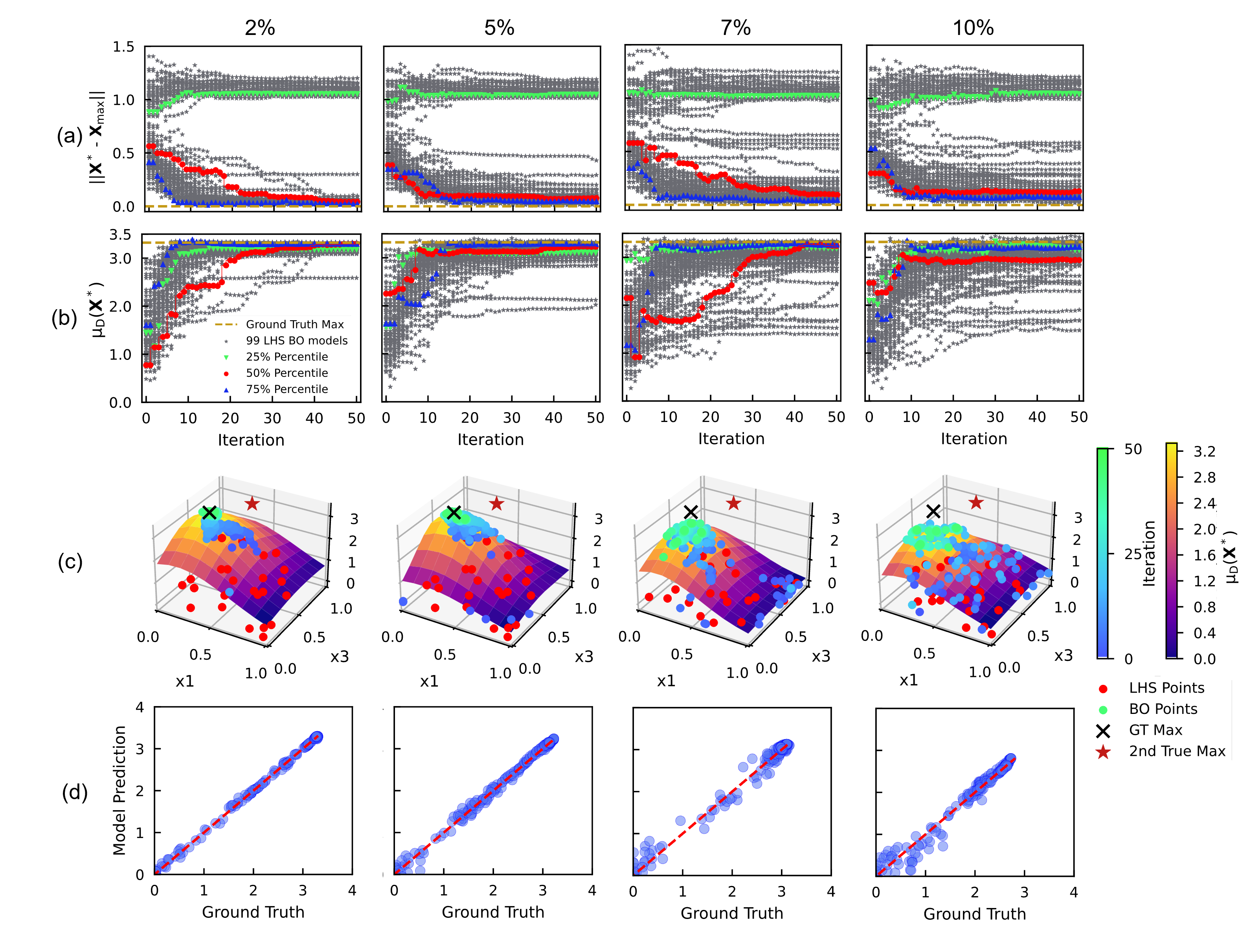}
    \caption{Learning curve of $||{\bf X}^* - {\bf X}_{\rm max} ||$ in top row (a) and $\mu_{\rm D} ({\bf X}^*)$ in the second row (b) of the Hartmann function using LP as the batch-picking method and EI as the acquisition function for noise levels of 2\% ($\xi$ = 0.005), 5\% ($\xi$ = 0.1), 7\% ($\xi$ = 0), and 10\% ($\xi$ = 0) from left to right. For visualization, (c) 3D representations and (d) parity plots of the 50$^{\rm th}$ percentile BO models for the same noise levels. In the 3D representations, x1 and x2 are projected with 224 learning points, including 24 initial LHS points (red) and 50 iterations (from blue to green) of BO points.}
    \label{fig:8}
\end{figure*}

Fig.~\ref{fig:8} (a) and (b) show the learning curves of the Hartmann function, using the EI acquisition policy, for noise levels of 2\%, 5\%, 7\%, and 10\%. At 2\% noise, almost all samplings reach either ${\bf X}_{\rm max}$ or ${\bf X}_{\rm max,2}$; only one LHS sampling results in ${\rm \langle IR}({\bf X})\rangle \approx 0.3$ and $\mu_{\rm D} ({\bf X}^*)<3$ indicating it is neither close to ${\bf X}_{\rm max}$ nor ${\bf X}_{\rm max,2}$. The spread on ${\rm \langle IR}({\bf X})\rangle$ increases with the noise but the 50$^{\rm th}$ percentile sampling still approaches the ${\bf X}_{\rm max}$ even at 10\% noise, distinctly different from the Ackley case (Fig.~\ref{fig:7}). 

Fig.~\ref{fig:8}(c) illustrates the GPR model of the 50$^{\rm th}$ percentile sampling and Fig.~\ref{fig:8}(d) shows the corresponding parity plots. Since the Hartmann function is not symmetric, the projections onto the other pairs of input variables are shown in Fig.~\textbf{S17}-~\textbf{S20}. At low noise (2\% and 5\%), almost all of the ${\bf X}$ are in the approximate vicinity of ${\bf X}_{\rm max}$. At 7\% noise, we observe that a region with $x1\approx 1$ and $x2 \in[0.5,1]$ is explored initially but finally ${\bf X}_{\rm max}$ is found. At 10\% noise many points are sampled with $x2 \in [0,0.5]$ but ${\bf X}_{\rm max}$ is found, and the landscape still visually resembles the Hartmann landscape, albeit with a reduced height at the maximum. The GNV hyperparameters for the noisy cases (Fig.~\textbf{S21}) reflect the increasing noise level. BO still converges with a 15\% noise level (Fig.~\textbf{S22}). The performance of BO degrades with noise with Hartmann test function but, in contrast with Ackley test function, BO remains functional for Hartmann up to 15\% noise.

We investigated the optimal exploration hyperparameters for both UCB and EI acquisition functions as a function of noise, based on the performance metric ${\rm 
\langle IR}({\bf X})\rangle$ (Table ~\textbf{S12} and ~\textbf{S13}). Trends similar to the Ackley test function are found for Hartmann, albeit Hartmann test function is less sensitive to the choice. The smallest ${\rm \langle IR}({\bf X})\rangle$-yielding values are used for Fig.\ref{fig:8} (EI with $\xi=0.005$, $0.1$, $0$, $0$ for 2\%, 5\%, 7\%, and 10\% noise, respectively). Table ~\textbf{S14} highlights the best-performing acquisition function and hyperparameter for Hartmann test function at each noise level. EI is the best acquisition function for five of the noise conditions whereas UCB is the best for six of the noise conditions with only small differences between each. This indicates that for the Hartmann landscape, UCB and EI have similar BO performance.

Fig. \ref{fig:9} and in Movie~(6) (see supplementary) show the full iterative progress of the BO of the Hartmann function with 5\% noise using the EI acquisition policy ($\xi$ = 0.1). Initially, the maximum of the GPR landscape is not located near ${\bf X}_{\rm max}$. However, at iteration 32, the optimum has been identified. In comparison, in the noise-free case (Fig. ~\textbf{S8}), ${\bf X}_{\rm max}$ and $H({\bf X}_{\rm max})$ are identified after only 5 iterations.
\begin{figure*}[h!]
    \centering
    \includegraphics[width=6in,keepaspectratio]{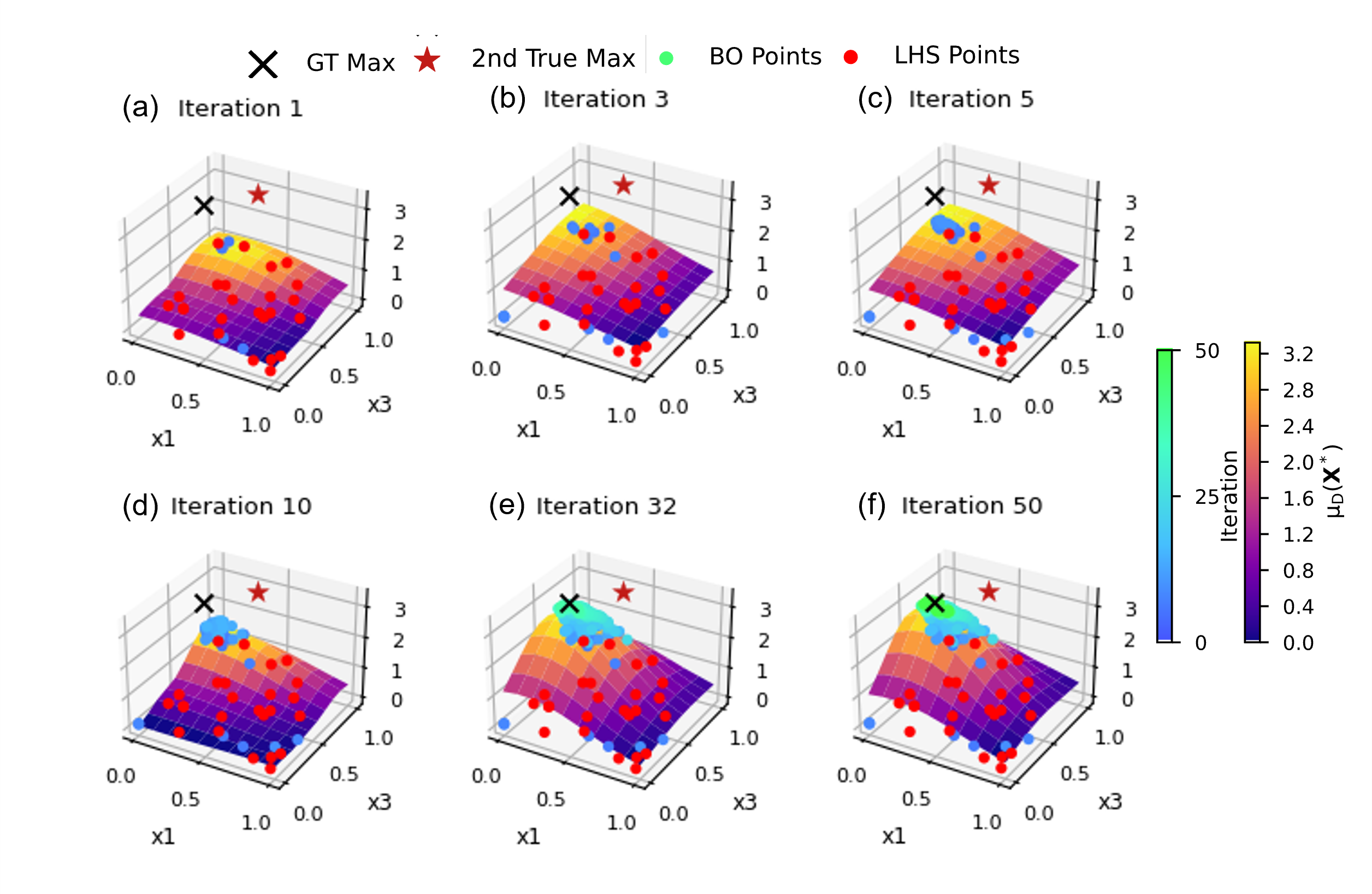}
    \caption{(a)-(f) 3D representations of 50$^{\rm th}$ percentile BO models (EI, $\xi$= 0.1) of the Hartmann test function with 5\% noise at different iterations, where x1 and x3 are shown with 224 learning points (including 24 initial LHS points in red). Black cross marks ${\bf X}_{\rm max}$ and red stars marks ${\bf X}_{\rm max,2}$}
    \label{fig:9}
\end{figure*}

\begin{figure}[h!]
    \centering
    \includegraphics[width=0.8\linewidth,keepaspectratio]{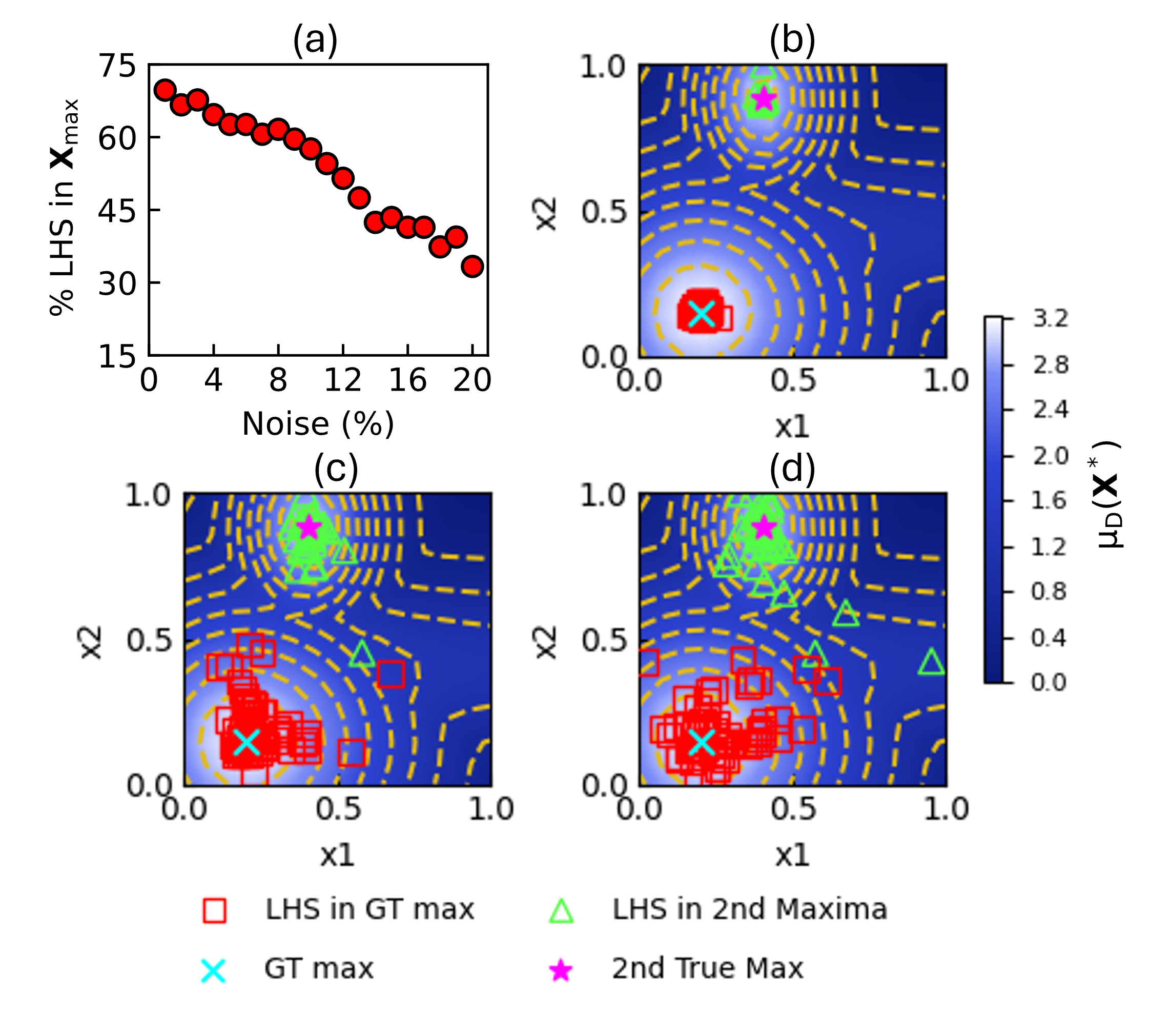}
    \caption{For Hartmann function, (a) percentage of data in ${\bf X}_{\rm max}$ as a function of noise levels, and projection of ${{\bf X}^{50}}^*$ for all 99 LHS samplings onto the GT 2D heat maps with noise level equal to (b) 0\%,(c) 10\%, and (d) 20\%. Red squares are represent ${{\bf X}^{50}}^*$ closer to ${\bf X}_{\rm max}$ and green triangles represent ${{\bf X}^{50}}^*$ closer to ${\bf X}_{\rm max,2}$ at the end of 50 iterations.}
    \label{fig:10}
\end{figure}

Fig. \ref{fig:10}(a) shows the fraction of LHS samplings where $||{\bf X}^{50}-{\bf X}_{\rm max}||<||{\bf X}^{50}-{\bf X}_{\rm max,2}||$. As noise increases, the fraction of LHS samplings that find ${\bf X}_{\rm max}$ decreases from 75\% to 30\%. The observation matches with Fig.~\ref{fig:8}, in which we studied noise up to 10\% and observed that the 50th percentile still find ${\bf X}_{\rm max}$. However, further increase in noise results in more than 50\% of the LHS samplings ending up closer to ${\bf X}_{\rm max,2}$ compared to ${\bf X}_{\rm max}$. This indicates that for a landscape with almost degenerate maxima, at high noise, the BO is not able to distinguish between the two. This could arise from the area of the convex peak of the second maximum being large compared to the first, as well as the relative locations of the two optima. Whenever the noise level is higher than the difference in the values of the true global optimum and other competing optima, BO has only little evidence for determining which one of the optima is better. However, the ability of BO to find an optimum diminishes more slowly with the increasing noise than the ability to find the global optimum.

Locations ${\bf X}^{50}$ are illustrated in Fig.~\ref{fig:10}(b-d) for 0\%, 10\%, and 20\% noise levels. Without noise, all samplings result in ${\bf X}^{50}$ are clearly near either ${\bf X}_{\rm max}$ or ${\bf X}_{\rm max,2}$. But at 10\% noise, a few samplings end up in between the two locations and are attributed to the first or second maximum only through the criterion $||{\bf X}^{50}-{\bf X}_{\rm max}||<||{\bf X}^{50}-{\bf X}_{\rm max,2}||$. The number of "stray" samplings increases when noise increases to 20\%.

\subsubsection{Simulating Noise Amplitude to Better Resemble Experiments}

Thus far in the benchmarking studies, we have incorporated noise as a percentage of ${\rm Max}(y_{\rm GT})$ for both test functions, as commonly done in the BO literature.\cite{D1ME00154J,wu,Picheny,letham2018constrainedbayesianoptimizationnoisy} Noise is thus incorporated as a Gaussian distribution with a zero mean and the standard deviation of the specified noise level (Eq.~($\eqref{8}$)), which is added to the objective function value, $f({\bf X})$, according to Eq.~($\eqref{9}$).

While adding noise in this fashion is convenient in simulations, experimentally, noise is referenced to signals represented by the signal-to-noise ratio (SNR), which does not have a direct correspondence in the BO approach. First, when running simulated BO, noise incorporation proportionally to $\rm Max(y_{GT})$ can significantly overestimate the general noise level.\cite{daulton2022} Secondly, since the benchmarks are utilized for making informed decisions for BO settings of future experiments, the noise level of the experiment to be used in the benchmarks is estimated based on repetitions of samples on a few points of the search space. In this case, the selected points are likely not to represent the true maximum since otherwise the experimental BO campaign would hardly be initiated in the first place. At the same time, noise proportional to $\rm Max(y_{GT})$ is relatively more detrimental for heterogeneous functions such as the Ackley than for smooth domains like the Hartmann, as seen in Figs. \ref{fig:7} and \ref{fig:8}, which complicates the interpretation of the consequences. In experimental setups with multi-dimensional input parameters and complex objective functions, the over-estimation of the noise is generally considered safe. However, as a downside, the over-estimation of noise can lead to performing more experimental evaluations than necessary, increasing time and cost.
\begin{figure}[h!]
    \centering
    \includegraphics[width=0.8\linewidth,keepaspectratio]{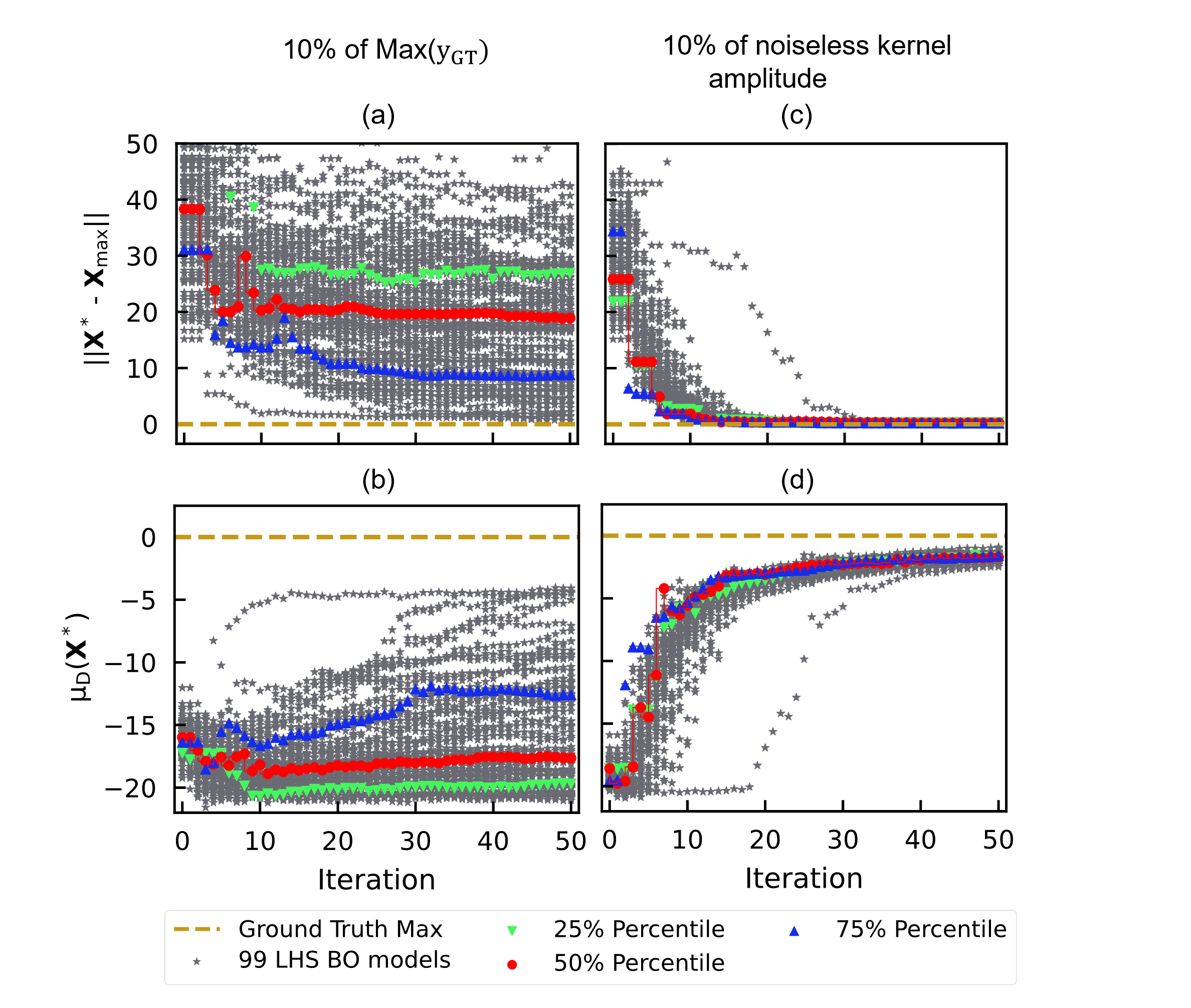}
    \caption{Ackley learning curve of $||{\bf X}^* - {\bf X}_{\rm max} ||$ (top row) and $\mu_{\rm D} ({\bf X}^*)$ (bottom row) for 10\% noise: (a) and (b) of $\rm Max(y_{GT})$ and (c) and (d) of noiseless kernel amplitude. UCB is used as the acquisition policy. }
    \label{fig:11}
\end{figure}

Therefore, we considered other options to incorporate noise in the benchmarking studies. We argue that the kernel amplitudes under noiseless conditions can be taken as the signal level. Thus, a more physical way to set the noise level is setting noise proportionally to the noiseless kernel amplitude, as expressed in Eq.~$\eqref{10}$, rather than $\rm Max(y_{GT})$ (Eq.~$\eqref{9}$). This choice stemmed from the observation that the kernel amplitude plays a pivotal role in enhancing the precision of predictions for the objective function during the BO process.

Fig. \ref{fig:11} compares 10\% noise case incorporated in two noise frameworks for the Ackley function: proportional to $\rm Max(y_{GT})$ (a-b) and proportional to the noiseless kernel amplitude (c-d). The comparison for the Hartmann function is shown in Fig.~\textbf{S23}. The noiseless kernel amplitude is equal to 0.192 and 0.184 for Ackley and Hartmann, respectively. For the kernel amplitude noise framework (Fig. \ref{fig:11}(c-d)), all final BO models are well-optimized for the Ackley case. Conversely, for the $\rm Max(y_{GT})$ noise framework (Fig. \ref{fig:11}(a-b)), only around 25\% of the final BO models exhibit successful optimization, while the remaining fail to achieve satisfactory levels of optimization. The ${\rm \langle IR}({\bf X})\rangle$ value of the $\rm Max(y_{GT})$ is two orders of magnitude larger than that of  kernel amplitude noise frameworks: $31.7\times10^{-2}$ and $0.38\times10^{-2}$, reflecting the effect of noise over-estimation on BO outcome.

In the Hartmann case (Fig.~\textbf{S23}) with the kernel amplitude noise framework (Fig..~\textbf{S23}(c-d)), again all final BO models converge to ${\bf X}_{\rm max}$ or ${\bf X}_{\rm max,2}$. Conversely, for the $\rm Max(y_{GT})$ noise framework (Fig..~\textbf{S23}(a-b)), only around 40-50\% of the final BO models attain successful optimization, while the rest end up in non-optimal regions. However, the noise over-estimation from setting noise level using $\rm Max(y_{GT})$ has a more detrimental effect on heterogeneous problems than non-heterogeneous problems. Adapting the kernel noise framework mitigates this problem and result in good convergence for both functions.

\subsubsection{Effects of Noise Depend on the Problem Landscape}

\begin{figure*}[h!]
    \centering
    \includegraphics[width=0.65\linewidth,keepaspectratio]{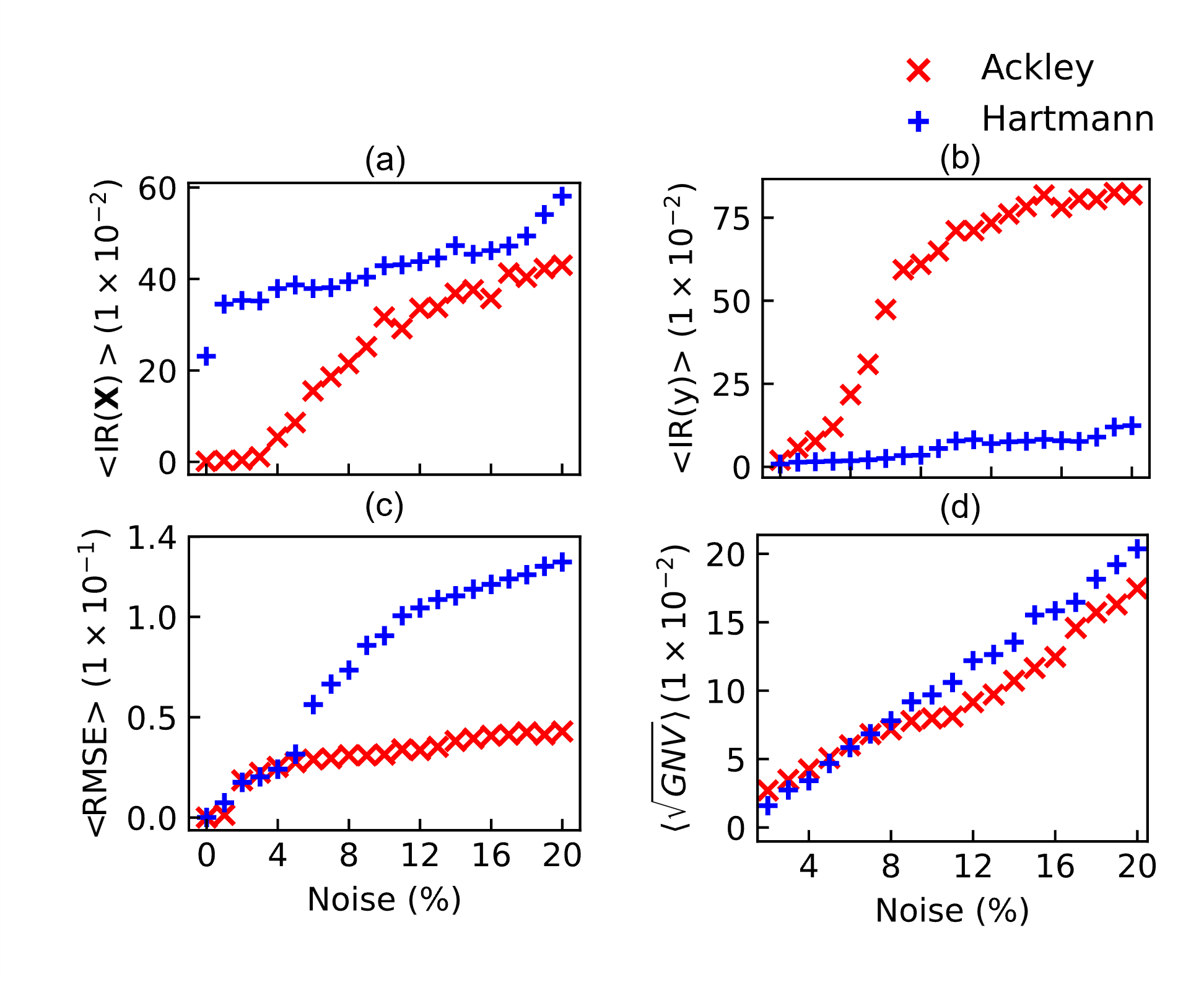}
    \caption{Noise dependence of (a)normalized $\rm \langle IR(\bf X)\rangle$, (b) normalized $\rm \langle IR(y)\rangle$, and (c) normalized $\rm \langle RMSE \rangle$, and (d) $\rm \langle \sqrt{GNV} \rangle$. All averaged over 99 LHS samplings. For all plots, Ackley test function results are represented by red x's and Hartmann results are blue +'s. EI is used as the acquisition function and LP is used as batch-picking method.}
    \label{fig:12}
\end{figure*}

Fig.~\ref{fig:12} (a) and (b) depict ${\rm \langle IR}({\bf X})\rangle$ and ${\rm \langle IR}(y)\rangle$ (normalized by the GT function $\bf X$ and $y$ ranges) as a function of noise level. For Hartmann, ${\rm \langle IR}({\bf X})\rangle \approx 0.2$ without noise and increases to ${\approx 0.35}$ when a small amount of noise is introduced. For Ackley, ${{\rm \langle IR}({\bf X})\rangle}\approx 0$ up to 4\% noise after which the ${\rm \langle IR}({\bf X})\rangle$ increases up to 0.4 for 20\% noise. The large ${\rm \langle IR}({\bf X})\rangle$ value of Hartmann even without noise is caused by some BO models ending up in ${\bf X}_{\rm max,2}$. The strong dependence of ${\rm \langle IR}({\bf X})\rangle$ for Ackley is the result of noise causing the BO to fail completely to find the optimum. Inspecting Fig.~\ref{fig:12}b reveals that ${\rm \langle IR}(y)\rangle$ increases rapidly for the Ackley function and reaches 0.5 at 8\% noise already due to many of the GPR models not being able to replicate the Ackley peak (see Fig.~\ref{fig:7}c). For the Hartmann function, even at 20\%, $\langle{\rm IR}(y)\rangle \approx 0.1$. Thus, it is more challenging to fit heterogeneous-type objectives.

Fig.~\ref{fig:12}c shows that $\rm \langle RMSE \rangle$ values (averaged from 99 LHS BO models' RMSE's) for both functions increase with noise levels, with the two being approximately the same for $\langle$ 5\% noise. At higher noise levels, $\rm \langle RMSE \rangle$ increases much more rapidly for Hartmann compared to Ackley. At first glance, this result is anti-intuitive since the GPR model at 10\% noise level does not resemble the Ackley function at all (Fig.~\ref{fig:7}c), yet the RMSE value is lower than the Hartmann case where the GT landscape is clearly visible (Fig.~\ref{fig:8}c).

This peculiar result is the direct consequence of the Ackley peak occupies only a miniscule fraction of the hypervolume. With the increasing noise, the GPR models completely missed the Ackley optimum region, but instead explore the plateau region where the GT values are small. The small $\rm \langle RMSE \rangle$ between the GT and GPR prediction indicates that BO correctly models the plateau region, and the large ${\rm \langle IR}(y)\rangle$ comes from missing the peak. Finally, the square root of the Gaussian noise variances in the final models averaged over all 99 LHS samplings, $\rm \langle \sqrt{GNV}\rangle$, is shown in Fig.~\ref{fig:12}d as function of noise. Up to 9\% noise, $\rm \langle \sqrt{GNV}\rangle$ values are similar for both the Hartmann and Ackley functions, but beyond 9\%, Ackley's $\rm \langle \sqrt{GNV}\rangle$ becomes slightly lower. From 10\% noise onwards, fitting the GPR GNV hyperparameter becomes more challenging for Ackley, while Hartmann's $\rm \langle \sqrt{GNV}\rangle$ shows a stronger correlation with increasing noise levels, as expected.

\section{Conclusion}\label{sec4}

Bayesian optimization (BO) is a machine learning method for effective optimization tasks; in experimental materials science, BO is useful for problems such as composition optimizations of mixed materials to achieve a desired but highly unusual property and process parameter optimizations to improve yield. In this work, we investigated high-dimensional BO --- representing multi-variable optimization problems --- and its relevant simulation and design choices in scenarios that involve non-negligible noise and performing work in batches, {\it i.e.}, under conditions that are typical in materials experiments but not common in machine learning literature on BO. 

We applied batch BO to the 6D Ackley function, representative of needle-in-a-haystack search problems such as searching for auxetic materials with a negative Poisson's ratio,\cite{deJong2015} materials with high thermoelectric figure of meric (ZT),\cite{Hinterleitner} and high-entropy alloys that exhibit both high strength and high ductility,\cite{Li2016} and the 6D Hartmann function, representative of a problem with a false maximum that are nearly degenerate with the global maximum, exemplified by optimizing deposition parameters for perovskite solar cells,\cite{Liu2022} searching for stable perovskite composition,\cite{Sun2021}, synthesizing silver nanoparticle by microfluidics,\cite{mekki-berrada2021two} and identifying print parameters for enhanced output quality.\cite{deneault2021toward}  

We developed a benchmark and visualization framework for planning and tracking high-dimensional BO, because problems above three dimensions are difficult to visualize and thus, controlling the high-dimensional optimizations is commonly challenged by difficulties to extract relevant information on its progress in a timely manner. We propose visualizing the optimization progress with (1) learning curves of both design variables $\bf X$ and objective variable $y$, (2) the evolution of hyperparameters, (3) 3D projections of the final Gaussian process regression (GPR) surrogate models as a function of BO iterations, and (4) parity plots representing the match of the experiment data and the surrogate model predictions.

We found that in the absence of noise, BO is not challenged by the needle-in-a-haystack type Ackley ground truth function  and efficiently finds the input values associated with the global maximum $({\bf X}_{\rm max})$. In contrast, the Hartmann function sets a tougher challenge due to its local optimum $({\bf X}_{\rm max,2})$ close to ${\bf X}_{\rm max}$, with 30\% of the LHS samplings ending up at ${\bf X}_{\rm max,2}$. This is reflected in the large instantaneous regret (${\rm \langle IR}({\bf X})\rangle$) value for Hartmann even without noise. In the absence of noise, the UCB acquisition function with exploration hyperparameter $\beta=1$ yields the best performance compared to other values of $\beta$ or acquisition function EI with any tested value of its exploration hyperparameter for both Ackley and Hartmann when evaluated according to ${\rm \langle IR}({\bf X})\rangle$, ${\rm \langle IR}(y)\rangle$ and cumulative regret ${\rm \langle CR}({\bf X})\rangle$. Only for ${\rm \langle CR}(y)\rangle$ in the Hartmann function does EI slightly edge out UCB.

We demonstrated that in noisy optimization scenarios as in typical materials science experiments, the progress of BO should be monitored via the predicted values of the objective function (the optimal posterior mean of the surrogate model, $\mu_{\rm D} ({\bf X}^*)$) rather than the best objective value collected (${\rm Max}(y)$). This is a more robust way to track the BO convergence because $\mu_{\rm D} ({\bf X}^*)$ is less susceptible to outliers. We showed that BO performance is strongly degraded for the Ackley function by noise, with BO not able to converge to the ground truth maximum when 10\% noise is present. For the Hartmann function, we also observed degradation but convergence to the optimum is maintained up to 15\% noise even though the convergence is increasingly distracted by the local optimum of the objective function ${\bf X}_{\rm max,2}$. This difference is reflected in our benchmark metrics: ${\rm \langle IR}({\bf X})\rangle$) for the Hartmann test function increases relatively slowly with the increasing noise level, while ${\rm \langle IR}({\bf X})\rangle$ for Ackley is highly dependent on the noise level, increasing drastically with noise above 4\% level. These observations show that the specifics of problem landscape, {\it i.e.}, the sensitivity of objectives on input variables, affect the BO performance in the presence of noise in experiments, and highlight the importance of domain expertise when adopting machine learning approaches for materials science research. 

As another factor relevant to planning experimental optimization campaigns in materials science, we showed that simulating noise with respect to the scale of the test function (here $\rm Max(y_{GT})$), as commonly done in the BO machine learning literature, might significantly over-estimate the noise in experiments. Fitting a GPR model on the objective and simulating noise with respect to the noiseless kernel amplitude of the same model likely represents more accurately the signal-to-noise ratio in experiments. In the latter case, BO is still able to optimize the needle-in-a-haystack Ackley function at 10\% noise whereas in the former case it is not. This highlights the importance of simulating noise in a realistic way prior to experiments, to be able to evaluate the required experimental budget and the feasibility of the optimization campaign correctly.

This work addresses many important, but often-neglected, issues, such as which acquisition function to use or how to decide what value of exploration hyperparameter to use, in multi-variable optimization encountered in materials science research. In addition to testing BO under noisy experimental conditions, we introduce visualization tools and performance metrics to monitor optimization progress. 
This lays the ground work for moving BO from controlled studies toward robust, real-world adoption, bridging the gap between theory and experimental practice in materials research.

\section*{Funding}
This work is supported by National Science Foundation CMMI-2135203. A.T. acknowledges funding from the European Union’s Horizon 2020 research and innovation programme under the Marie Skłodowska-Curie grant agreement No 01059891. J.W.P.H. and T.B. acknowledge the support from the Simons Foundation Pivot Fellowship. J.W.P.H. acknowledges the support of the Texas Instruments Distinguished Chair in Nanoelectronics.

\section*{Acknowledgements}
We thank M. Lee, R. Garnett, B. Das, K. Brown, and K. Snapp for helpful discussions. We acknowledge the Texas Advanced Computing Center (TACC) at the University of Texas at Austin for providing the high-performance computing resources that have contributed to the research results reported within this paper. URL: 
\url{http://www.tacc.utexas.edu}.

\section*{Author Contributions}
I.M. - data curation, formal analysis, investigation, software, validation, visualization, writing - original draft, writing - review \& editing. A.T. - methodology, validation, writing - review \& editing. A.E. and A.S. - software. T.B. - methodology. W.V. - resources, software, supervision, writing - review \& editing. J.W.P.H. - conceptualization, funding acquisition, methodology, project administration, resources, supervision, writing - review \& editing.

\section*{Conflicts of Interest}
All authors declare no conflicts of interest.

\section*{Data and Code Availability}
Python codes and data reported in the main text and supplementary materials are available on GitHub:  
\href{https://github.com/UTD-Hsu-Lab/Noisy-BO}{\nolinkurl{https://github.com/UTD-Hsu-Lab/Noisy-BO}}.  
During the review, reviewers can access the same information via  
\href{https://utdallas.box.com/s/kb3u37odpok0bfgs93ijkb3mnmdc0h6n}%
{\nolinkurl{https://utdallas.box.com/s/l59ogf1mu5va29xxyoi7dr3i9nplvrf8}}.

\section*{Supplementary Information}

The supplementary materials accompanying this paper include

1. A PDF file with mathematical description and complete 3D visualization of the test functions, performance metrics for different exploration hyperparameter values for both acquisition functions and test functions, plots of kernel hyperparametr evolution during the optimization for both test functions, Ackley results with optimum location not at the center of the hypervolume, 3D visualization of Hartmann optimization results, comparisons between UCB and EI learning results for both test functions, comparisons of local penalization, Kriging believer, and constant liar for both acquisition functions and test functions, optimal acquisition function and hyperparameter value for noise levels from 0 to 20 \% for both functions using ${\rm \langle IR}({\bf X})\rangle$ as the metric, and additional results for different noise levels. 

2. Six animated videos illustrating the optimization process where the iteration progress is represented through the time evolution: (1) the $\rm 50^{th}$ percentile GPR model projected to x1-x2 plane for the noiseless Ackley function; (2) the $\rm 50^{th}$ percentile GPR model projected to x1-x3 plane for the noiseless Hartmann function; (3) the $\rm 25^{th}$ percentile GPT model (left) and (4) $\rm 75^{th}$ percentile GPR model projected to x1-x3 plane for the noiseless Hartmann function; (5) the $\rm 50^{th}$ percentile GPR model projected to x1-x2 plane for the Ackley function with 5 noise; (6) the $\rm 50^{th}$ percentile GPR model projected to x1-x3 plane for the Hartmann function with 5 noise.




\end{document}